\documentclass{article}

\usepackage{iclr2027_conference,times}
\usepackage{amsmath}

\usepackage[utf8]{inputenc}
\usepackage[T1]{fontenc}
\usepackage{xcolor}
\definecolor{iccvblue}{rgb}{0.21,0.49,0.74}
\usepackage[pagebackref,
            breaklinks,
            colorlinks=true,
            linkcolor=iccvblue,
            citecolor=iccvblue,
            urlcolor=iccvblue]{hyperref}
\usepackage{url}
\usepackage{booktabs}
\usepackage{amsfonts}
\usepackage{microtype}
\usepackage[most]{tcolorbox}
\usepackage{listings}
\tcbuselibrary{listings}
\usepackage{wrapfig}
\usepackage{subcaption}

\newtcolorbox{takeaway}[1]{%
  enhanced,
  colback=gray!8,
  colframe=gray!55,
  arc=2mm,
  boxrule=0.5pt,
  left=6pt, right=6pt, top=8pt, bottom=6pt,
  attach boxed title to top left={xshift=4mm,yshift=-2.8mm},
  boxed title style={
    colback=gray!70,
    colframe=gray!70,
    arc=1mm,
    boxrule=0pt,
    left=4pt, right=4pt, top=1.5pt, bottom=1.5pt,
  },
  fonttitle=\scriptsize\bfseries\scshape\color{white},
  title={Takeaway~#1},
}

\newtheorem{lemma}{Lemma}
\newtheorem{proposition}{Proposition}
\newtheorem{proof}{Proof}

\usepackage{etoolbox}
\makeatletter
\g@addto@macro\normalsize{%
  \setlength{\abovedisplayskip}{2pt plus 1pt minus 1pt}%
  \setlength{\belowdisplayskip}{-3pt plus 1pt minus 1pt}%
  \setlength{\abovedisplayshortskip}{2pt plus 1pt minus 1pt}%
  \setlength{\belowdisplayshortskip}{-3pt plus 1pt minus 1pt}%
}
\g@addto@macro\small{%
  \setlength{\abovedisplayskip}{2pt plus 1pt minus 1pt}%
  \setlength{\belowdisplayskip}{-4pt plus 1pt minus 1pt}%
  \setlength{\abovedisplayshortskip}{1pt plus 1pt minus 1pt}%
  \setlength{\belowdisplayshortskip}{-5pt plus 1pt minus 1pt}%
}
\g@addto@macro\footnotesize{%
  \setlength{\abovedisplayskip}{2pt plus 1pt minus 1pt}%
  \setlength{\belowdisplayskip}{-4pt plus 1pt minus 1pt}%
  \setlength{\abovedisplayshortskip}{1pt plus 1pt minus 1pt}%
  \setlength{\belowdisplayshortskip}{-5pt plus 1pt minus 1pt}%
}
\makeatother
\AtBeginDocument{\normalsize}
\AtBeginEnvironment{proposition}{\vspace{-6pt}}
\AtEndEnvironment{proposition}{\vspace{-6pt}}
\AtBeginEnvironment{lemma}{\vspace{-6pt}}
\AtEndEnvironment{lemma}{\vspace{-6pt}}

\newcommand{\E}{\mathbb{E}}
\newcommand{\Prob}{\mathbb{P}}
\newcommand{\1}{\mathbf{1}}
\newcommand{\calS}{\mathcal{S}}
\newcommand{\calA}{\mathcal{A}}
\newcommand{\calY}{\mathcal{Y}}
\newcommand{\ystar}{\mathcal{Y}^\star}

\newcommand\mypar[1]{\par\vspace{-0.0mm}\noindent\textbf{#1}\;\;}

\title{Not All Reasoning Generalizes: A Case Study on Learning Generalizable Reasoning Strategies in RLVR}

\author{Ziqi Pang \quad Yu-Xiong Wang \\
University of Illinois Urbana-Champaign \\
\texttt{\{ziqip2,yxw\}@illinois.edu}
}

\iclrpreprint
\begin{document}

\maketitle

\begin{abstract}
Reinforcement learning with verifiable rewards (RLVR) can enable generalizable improvement for large language model (LLM) reasoning. This paper studies how to incentivize such generalization via \textbf{reasoning strategies}. We study this question with a controllable equation-solving framework: finding multiple inputs $x$ for equations $f(x)=y$, where $f$ is a non-bijective string function. This setting makes reasoning strategies observable while supporting a clear definition of easy-to-hard out-of-distribution (OOD) shifts, \emph{e.g.}, more solutions, longer $y$, etc. We uncover a subtle RLVR failure: it teaches the model to solve in-distribution tasks with high accuracy and structured step-by-step derivation, yet under OOD shifts, the same model abandons derivation and collapses to ineffective brute-force enumeration, causing sharp accuracy drops. This observation reveals the deeper challenges of RLVR generalization as it is neither \emph{incapable base model} nor \emph{reward hacking}, since the model succeeds in-distribution with the correct reasoning strategy. The key to mitigating this failure is an indirect intervention that surprisingly helps: co-training with bijective inversion tasks. Although these bijective tasks are actually \emph{not uniformly harder} and even \emph{further} from the non-bijective OOD evaluation, they make derivation more certain and substantially restore the correct strategy and accuracy under OOD shifts. This effect further motivates our theoretical analysis: outcome-only RLVR can shape reasoning processes through implicitly crediting the effective reasoning strategies more, even without process supervision. To conclude, our analysis framework and findings systematically reveal the crucial role and incentivization of reasoning strategies in RLVR generalization under OOD shifts. Code is available at \url{https://github.com/ziqipang/rl_meta_thinking}.
\end{abstract}

\section{Introduction}
\label{sec:introduction}

Applying reinforcement learning with verifiable reward (RLVR) to large language models (LLMs) has shown surprising generalization capabilities. Rather than imitating fixed traces, RLVR incentivizes \emph{meta-level reasoning strategies} that can transfer to out-of-distribution (OOD) tasks. For instance, \citet{yuan2025fromfxgx,liu2025spiral} show that general reasoning strategies like composing and case analysis are shown to be transferable from the training tasks to novel domains~\citep{sun2025grokking}. Such positive cases, however, cannot establish transfer as a general rule. We ask the complementary question: \emph{when and why does RLVR generalization fail?}

\begin{figure}[t]
  \centering
  \includegraphics[width=0.99\linewidth]{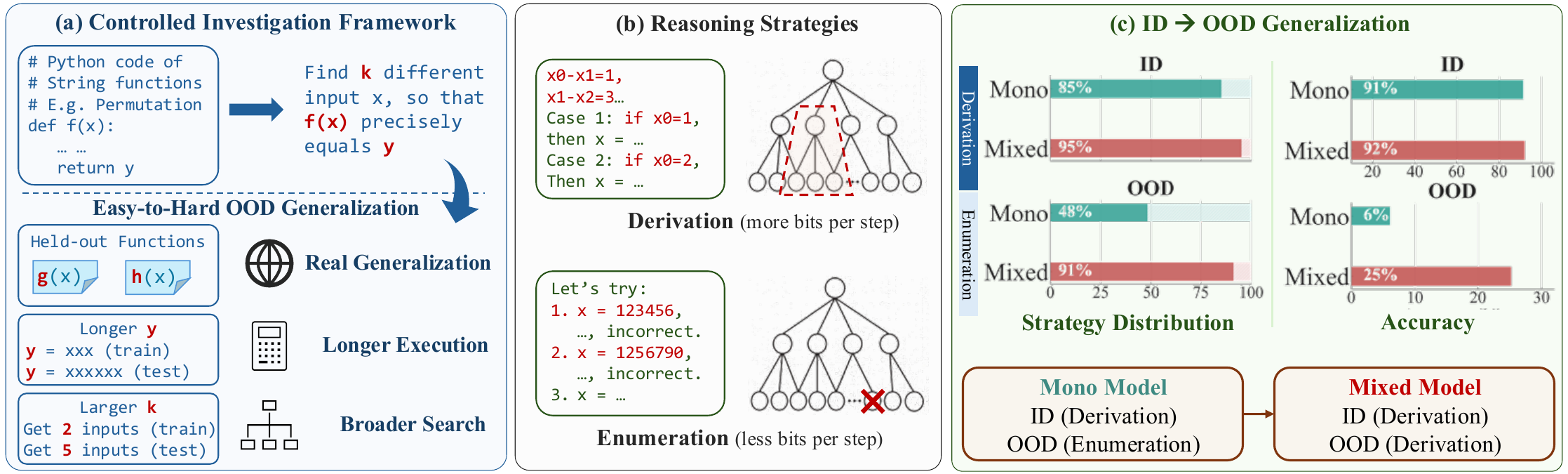}\vspace{-2mm}
  \caption{\textbf{(a)} To clearly define OOD generalization for LLM reasoning, we construct a controllable framework via solving string function equations, mirroring real-world math and coding. \textbf{(b)} We categorize two major reasoning strategies, the more effective \emph{Derivation} and or total brute-force \emph{Enumeration}. \textbf{(c)} We reveal how reasoning strategies reflect the challenges of OOD generalization. The \textsc{Mono} model learns derivation for ID cases, yet drastically collapses to enumeration for OOD. With \textsc{Mixed} better incentivizing the derivation strategy, it achieves much better OOD accuracy under similar ID performance. These form a systematic framework of understanding RLVR generalization.}\vspace{-4mm}
  \label{fig:teaser}
\end{figure}

However, we realize that such \emph{generalization} of reasoning strategies has not been thoroughly investigated. Previous studies mostly understand RLVR generalization from its relationship with the base models~\citep{yue2025does,yuan2025fromfxgx} without clearly categorizing individual reasoning strategies and analyzing their OOD generalization. To this end, we start by building a controllable framework that clearly defines OOD generalization and categorizes reasoning strategies. As in Fig.~\ref{fig:teaser}\textcolor{iccvblue}{a}, we construct a set of string functions $f(x)$ inspired by \citet{yuan2025fromfxgx}, and let the LLM address this \emph{inverse} problem: recovering multiple inputs $x$ such that $f(x)=y$ ($y$ is a provided string output). This task mirrors solving math equations and tracing bugs in coding. More importantly, we construct the functions $f(\cdot)$ to be non-bijective and require LLMs to output $k>1$ different solutions. This framework thus clearly splits in-distribution and OOD tasks, where the OOD tasks are built from held-out functions and made harder via longer strings $y$ and more required solutions $k$.

On this controllable framework, we observe that RLVR can reliably teach an LLM to \emph{perform reasoning effectively in-distribution}, showing reliable derivation strategies step-by-step and enumerating characters only when needed, as Fig.~\ref{fig:teaser}\textcolor{iccvblue}{b}. However, this model, named \textsc{Mono}, shows drastically different behaviors on harder OOD tasks: it abandons the strategy of derivation, and instead, relies on ineffective brute-force enumeration of all the digits simultaneously (Fig.~\ref{fig:teaser}\textcolor{iccvblue}{c}). Such an observation marks an important revelation: an LLM can \emph{learn to perform effective reasoning strategies in-distribution}, yet \emph{drastically falls back to worse strategies in OOD scenarios}. Such an observation is nuanced as it is not related to the commonly known RLVR failures like incapable base models or reward hacking, as the model is following valid reasoning traces to achieve high accuracy during training. Therefore, we are studying \emph{real} generalization challenges under this setting.

This diagnosis suggests that the key is not merely to improve training accuracy but to change which reasoning strategies RLVR incentivizes. We therefore introduce a seemingly indirect intervention for the task: co-training with bijective tasks that only asks for one solution $x$ for $f(x)=y$. The intuition is to decrease the search space in non-bijective functions and encourage step-by-step derivation with larger certainties. Importantly, this intervention is surprising because the bijective tasks are not \emph{uniformly harder} and are even \emph{further} away from the non-bijective OOD evaluation requiring multiple solutions. Nevertheless, the \textsc{Mixed} model with this intervention successfully restores the behavior of utilizing the derivation strategy and leads to significantly better OOD accuracy (Fig.~\ref{fig:teaser}\textcolor{iccvblue}{c}).

Such effectiveness inspires us to understand how better reasoning strategies can be incentivized from RLVR. As in Fig.~\ref{fig:teaser}\textcolor{iccvblue}{b}, we view the reasoning process as progressively shrinking the candidate solution space. This analogy clearly distinguishes the derivation and enumeration: derivation is more effective by structurally pruning multiple candidates in a single step, while enumeration can only contribute marginally by eliminating one wrong candidate. Therefore, the reasoning strategies like derivation can contribute more bits per step, and can more easily generalize to harder scenarios, \emph{i.e.}, those requiring more bits. By connecting this with RL algorithms theoretically, we show that even outcome-only RLVR can implicitly credit such effective reasoning processes without process supervision, explaining how reasoning strategies could emerge from RLVR. Beyond theory, our probing of the hidden states and reasoning trace audit also qualitatively shows how reasoning strategies could be a principled way to understand the generalization behaviors of LLMs in RLVR.

Our study presents a systematic analysis of RLVR generalization via reasoning strategies:

\vspace{-1.5mm}
\textbf{(i)} A controlled counter-example of RLVR teaching \emph{correct in-distribution reasoning strategies} that nevertheless \emph{change and fail under OOD shifts}: a subtle failure distinct from incapable base models.

\vspace{-1.5mm}
\textbf{(ii)} A principled intervention on reasoning strategies significantly enhances OOD generalization, which reveals the pivotal role of reasoning strategies in the transferable improvements of RLVR.

\vspace{-1.5mm}
\textbf{(iii)} The theoretical analysis explaining how more effective and generalizable reasoning strategies can be incentivized naturally via RLVR from a \emph{information per reasoning step} perspective.

\vspace{-1.5mm}
\textbf{(iv)} Reasoning trace inspection and probing of the model's hidden states qualitatively reveal how reasoning strategies are principled aspects to understand RLVR generalization.

Considering the importance of RLVR generalization, we hope this framework has provided a comprehensive yet nuanced understanding of LLM reasoning and generalization through synthetic and controllable data, with the same spirit as prior work that has inspired us~\citep{allenzhu2023physics1,ye2024physics,wang2024grokked,yin2025composable,yuan2025fromfxgx,zhang2025interplay}. 

\section{Related Work}
\label{sec:related_work}

\mypar{RLVR Generalization.}
RLVR has produced a growing set of positive generalization results: such as general reasoning and agentic capabilities enhancement after math/coding-oriented RL~\citep{shao2024deepseekmath,guo2025deepseekr1,kimiteam2025kimi}, carefully structured RL expands benchmark coverage~\citep{liu2025prorl,sun2025omega,sun2025grokking}, synthetic strings showing compositional expansion~\citep{yuan2025fromfxgx}, and game/puzzles improve to reasoning benchmarks~\citep{liu2025spiral,zhao2025absolutezero,xie2025play,tong2025gamerl}. Meanwhile, some studies argue that RLVR gains can also be re-distribution of the base model behaviors or spurious observations~\citep{yue2025does,wu2025invisible,liu2025drgrpo,zhao2025echo,gandhi2025cognitive,shao2025spurious,wu2025memorization,wu2025mirage}.
Our work, however, is complementary to both sides of this debate. Instead, we show a strategy-level and nuanced inspection of \emph{how} RLVR impacts the reasoning strategies of LLMs and the information effectiveness for a strategy to generalize to harder OOD domains. Specifically, a most crucial discovery is that RLVR can teach a model to solve the in-distribution task with the intended reasoning strategy, yet that strategy can collapse into a brittle alternative on OOD domains, which is different from either incapable-base-model failures or reward-hacking artifacts.

\mypar{RLVR Algorithms.} Representative LLM post-training algorithms include PPO-style RLHF~\citep{schulman2017ppo,ouyang2022training}, preference optimization like DPO~\citep{rafailov2023direct}, and outcome-based methods like GRPO~\citep{shao2024deepseekmath}. With the effectiveness of GRPO for math and coding, recent work has extended it through DAPO, Dr. GRPO, GSPO, and other variants or recipes that address MoE stability, length or token-level bias, sampling efficiency, etc~\citep{dapo,liu2025drgrpo,zheng2025gspo,hu2025openreasonerzero,zhu2025negative,wen2025implicit,liu2026luspo,zhang2025ar3po}. We mainly use the basic DAPO~\citep{dapo}, as our focus is the underlying meta-level reasoning behavior and generalization.

\mypar{Investigating LLMs via Controllable and Synthetic Data.} Synthetic data is a crucial tool to analyze the LLM behaviors reliably, especially represented by the physics of LLMs~\citep{allenzhu2023physics1,ye2024physics,ye2024physics22,allenzhu2023physics31,allenzhu2023physics32,allenzhu2024physics33}. In LLM reasoning, synthetic data has become essential for understanding generalization, since they provide clear boundaries: symbolic or length-controlled math variants~\citep{mirzadeh2024gsmsymbolic,zhou2025gsminfinite,liu2023tinygsm}, synthesized arithmetic probes~\citep{ye2024physics,wang2024grokked,yin2025composable}, and controlled agentic dynamics~\citep{wang2025ragen,wang2026ragen}. In addition, some studies also use causal interventions, circuit analyses, and probing to identify internal algorithms or encoded information~\citep{olsson2022incontext,meng2022locating,voita2020mdl,gurnee2023spacetime}. The work that inspires this work most is \citet{yuan2025fromfxgx}, which understands how meta-level reasoning skills like \emph{composing} could be generalized in RL. We adopt this controlled-data philosophy but use it for a different purpose: rather than using synthetic tasks only to avoid contamination or measure extrapolation, our inversion framework makes the model's \emph{reasoning strategy} observable and can reveal the nuances of RLVR generalization. With our synthetic framework, we uncover the collapse of RLVR in OOD scenarios, demonstrate principled interventions, and propose theoretical analysis.

\section{Research Framework}
\label{sec:design}

\subsection{Task Format and Principles}
\label{sec:design_principles}

\mypar{Principles.} Our investigation centers on \emph{which reasoning strategy} RLVR incentivizes, and whether the learned strategy generalizes from in-distribution training to OOD scenarios or quietly collapses. We therefore propose a research framework that surfaces the model's reasoning process and admits a controlled distribution shift between training and evaluation. The principles are: \textbf{(1)} \underline{\emph{Clean OOD Generalization.}} We adopt synthetic tasks following prior controlled LLM studies~\citep{yuan2025fromfxgx,sun2025grokking,chu2025sftrl,wang2025ragen,ye2024physics}, enabling a clean separation between the training and testing distributions for studying OOD generalization. \textbf{(2)} \underline{\emph{Explicit Reasoning Strategies.}} To diagnose reasoning strategies, the tasks must surface a small set of clearly identifiable strategies. The open-ended flexibility of real-world math or coding tasks works against this need for clean strategy categorization. \textbf{(3)} \underline{\emph{Reasoning Strategy Diversity.}} While the strategies must be categorizable, the framework must also allow multiple candidate strategies, so that generalization can be studied by observing which one RLVR selects. From this aspect, the homogeneous function-simulation tasks in \citet{yuan2025fromfxgx} fall short, as they only exercise the ``composing'' reasoning strategy and calculation of function outputs.

\mypar{The Inversion Task.} We instantiate these principles via an \emph{inversion} task on synthetic string functions, which reverses the direction of \citet{yuan2025fromfxgx}: given the source code of a function $f(\cdot)$ and a target output $y$, the LLM is asked to recover one or more digit-string inputs $x$ satisfying $f(x)=y$. This mirrors two real-world reasoning settings: solving math equations, where one reasons backward from a constraint to its solutions, and tracing program bugs, where one infers an upstream cause from an observed output. Crucially, the inversion task admits several reward-compatible strategies: analytical derivation, case analysis, and brute-force enumeration. It thus satisfies principles (2) and (3), while the next subsection addresses principle (1).

\subsection{Task Design}
\label{sec:task_design}

We turn the inversion task above into a concrete training-and-evaluation suite, including (i) a library of functions with a function-level train/held-out split, (ii) OOD tests that probe the learned strategy for easy-to-hard generalization, and (iii) evaluation metrics.

\mypar{Function Design for Held-out Training/Evaluation.}
We construct the training and testing sets to use distinct functions $f$, preventing the LLM from memorizing function-specific patterns. We therefore compose 24 non-bijective functions (many-to-one, so one output $y$ has multiple valid inputs $x$) (full list in Sec.~\ref{sec:appendix_function_lists}) and provide their definitions and Python code inside the prompt of every inversion task. Following \citet{yuan2025fromfxgx}, we split these functions into training and evaluation sets: 18 functions are used for RL training, and 6 strictly disjoint functions are held out for evaluation. This held-out slice forms the easiest test of generalization: same prompt interface, same input/output regime, only the underlying function is unseen. It serves as the in-distribution check that RLVR has learned to perform inversion. From this baseline, we then move to the \emph{easy-to-hard} OOD generalization tests below, which are our main probe of reasoning-strategy generalization. The training set contains 50k tasks, and each evaluation set contains 1{,}024 tasks.

\mypar{Easy-to-hard OOD Generalization.} As Fig.~\ref{fig:teaser}\textcolor{iccvblue}{a}, we design the in-domain and OOD tasks to differ along controlled axes. For the inversion task $f(x)=y$, the \emph{in-domain} scenarios train and evaluate the LLM on outputs $y$ of length 4--6 and require 2 candidate solutions per prompt. From this baseline, we design four easy-to-hard OOD axes: \textbf{(1)} \emph{More Answers} raises the number of requested solutions to a random integer between 3 and 6, demanding more computation and reliability to produce enough valid solutions. \textbf{(2)} \emph{Longer Outputs} increases the target output to length 6--10, lengthening the digit chain that must be inverted. \textbf{(3)} \emph{Longer Outputs \& More Answers} combines the two axes above. \textbf{(4)} \emph{Length Change} represents the largest shift: our in-distribution function family is designed so that $y$ and $x$ share the same length, but the Length-Change functions differ in the lengths of $y$ and $x$ (full list in Sec.~\ref{sec:appendix_function_lists}). Each OOD evaluation set contains 1{,}024 randomly generated tasks as above.

\mypar{Accuracy Scoring.} We compute accuracy based on whether the predicted solutions are correct with an outcome-based verifier. Each solution is scored semantically by executing the reference function and checking whether its output matches $y$. Let $k$ be the effective answer requirement and $m$ be the number of submitted solutions; accuracy is $\mathrm{acc}=\min(\text{correct\_unique}, k)/\max(k,m)$. This rewards answers that are both distinct and correct, and penalizes over-generation reward hacking.

\subsection{Training}
\label{sec:training_protocol}

We train Qwen2.5-7B-Instruct with GRPO~\citep{shao2024deepseekmath} and adopt the improvements from DAPO~\citep{dapo}, especially the filtering of groups with zero reward variance. This filtering is crucial both in the early stage of training, when the LLM is still developing its strategy with low accuracy, and in the late stage, when in-distribution accuracy has been high. We select this base model because our synthetic task lies on the boundary of its capabilities, requiring RL to incentivize correct reasoning strategies. Design details are in Sec.~\ref{sec:appendix_training_details}. Analysis of more base models in Sec.~\ref{sec:appendix_base_models}.

\section{Reasoning Strategy Collapse and Generalization}
\label{sec:main_observation}

This section reveals the collapse and generalization of reasoning strategies in OOD tasks. We first show how the reasoning strategies collapse in OOD tasks despite crrect in-distribution behaviors (Sec.~\ref{sec:finding1}). Then, we introduce incentivizing a better reasoning strategy for generalization (Sec.~\ref{sec:finding2}). 
\begin{figure}[t]
  \centering
  \begin{subfigure}[b]{\linewidth}
    \centering
    \includegraphics[width=\linewidth]{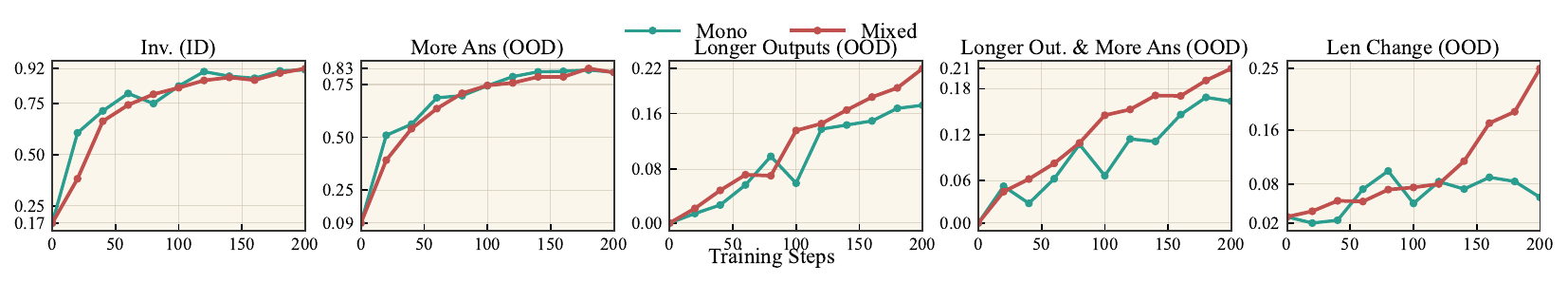}
    \vspace{-6mm}
    \caption{Evaluation Accuracy During RL}
    \label{fig:nbj_with_bj}
  \end{subfigure}\\[-2pt]
  \begin{subfigure}[b]{\linewidth}
    \centering
    \includegraphics[width=\linewidth]{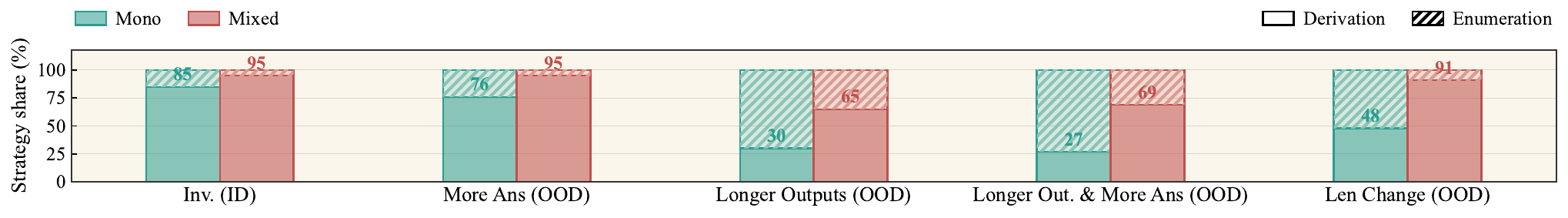}\vspace{-3mm}
    \caption{Reasoning Strategy Distribution}\vspace{-2pt}
    \label{fig:nbj_strategy}
  \end{subfigure}
  \vspace{-3mm}
  \caption{Reasoning strategy collapse and generalization in OOD tasks (Sec.~\ref{sec:main_observation}). The \textsc{Mono} model performs normally ID but collapses, which relies on brute-force enumeration OOD. In comparison, \textsc{Mixed} model shows similar ID performance, but surprisingly doubles analytical derivation reasoning on OOD and lead to better generalization. These reveal the challenges of reasoning strategy generalization and how to enhance the generalization of reasoning in RLVR.}\vspace{-4mm}
  \label{fig:main_obs}
\end{figure}

\subsection{\textsc{Mono} Model: Revealing OOD Collapse of Reasoning Strategies}
\label{sec:finding1}

Following Sec.~\ref{sec:design}, we train the LLM on the non-bijective function inversion tasks, \emph{i.e.}, solving $x$ given $f(x)=y$, and call this model \textbf{\textsc{Mono}}. By closely observing it on the in-distribution and the four easy-to-hard OOD sets, as in Fig.~\ref{fig:main_obs}, we observe the nuanced OOD reasoning strategy collapse.

\mypar{Reasoning Strategy Categorization.} Our synthetic tasks are specially constructed to understand the RLVR reasoning strategy generalization in a controllable way, as explained in Sec.~\ref{sec:design_principles}. So we 
\begin{wrapfigure}{r}{0.48\linewidth}
  \centering
  \includegraphics[width=\linewidth]{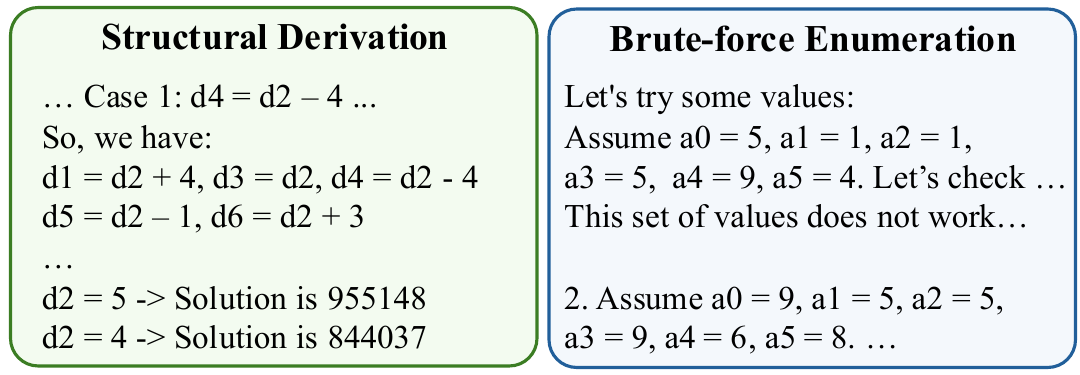}
  \vspace{-6mm}
  \caption{Examples of reasoning strategies.}
  \label{fig:strategy_examples}
\end{wrapfigure}
start from categorizing the reasoning strategy utilized by the \textsc{Mono} model. As in Fig.~\ref{fig:strategy_examples}, we can clearly attribute two primary strategies: \emph{derivation}, which structurally reverses $f$ and only conduct case-analysis when needed, and \emph{enumeration}, which directly guess all the digits all at once, being much less efficient. To make this binary classification revealing the traits of reasoning strategies, we treat purely brute-force enumeration as ``enumeration,'' and everything else as ``derivation.'' In practice, we employ an LLM judge empowered by Gemini-3.0-Flash to classify the reasoning traces (clarification of our setup, details, and verification of LLM judge in Sec.~\ref{sec:appendix_strategy_categorization}).

\mypar{From ID to OOD Collapse.} Fig.~\ref{fig:nbj_with_bj} tracks accuracy during RL on the ID set and the four OOD sets, and Fig.~\ref{fig:nbj_strategy} shows which strategies produce these responses. On in-distribution tasks, the \textsc{Mono} model learns the effective strategy of derivation in most cases and achieves high accuracy (as Fig.~\ref{fig:main_obs}, ID columns). However, its failure in OOD scenarios, \emph{e.g.}, the novel functions of lenchange, is not solely due to longer reasoning or more computation. Rather, its reasoning skill of derivation acquired disappears on OOD tasks and gives way to the inefficient brute-force enumeration, which proposes all the digits of $x$ simultaneously, without utilizing the structure of the function $f$. The majority of failures are caused by loops that enumerate all possible digits until reaching the token limits. Here, \emph{collapse} refers to the reasoning strategy: \textsc{Mono}'s derivation share drops from 85\% on ID to below 50\% on the three harder OOD sets. Accuracy, in contrast, degrades progressively as the shift grows: it stays above 80\% on More Answers, the mildest shift, and drops to 6\% on Length Change. 

Such an observation is surprising as it challenges RLVR generalization: \emph{the reasoning strategy already incentivized in RL fails to generalize despite identical task formats}. As mentioned in Sec.~\ref{sec:introduction} and Sec.~\ref{sec:design}, this observation focuses on the meta-level reasoning skills of LLMs and is subtle and distinct from prior work. First, it is not a \emph{base-model} limitation: the same model reaches $\approx{}90\%$ on in-distribution non-bijective inversion. Second, it is not \emph{reward hacking}: the LLM uses the correct and efficient derivation strategy for in-distribution evaluation rather than spurious shortcuts. We refer to this phenomenon as \emph{reasoning-strategy collapse}, which motivates our central question: which training signal enables RLVR to generalize a reasoning pattern to OOD scenarios?
\vspace{-1mm}
\begin{takeaway}{1}
Reasoning Strategy Collapse: Even though RLVR has incentivized the correct reasoning strategies for in-distribution tasks, the LLM might drastically change to ineffective strategies on OOD tasks despite identical task formats, undermining the generalization of RLVR.
\end{takeaway}
\vspace{-1mm}

\subsection{\textsc{Mixed} Model: Enhancing OOD Generalization via Bijective Inversion}
\label{sec:finding2}

The \emph{derivation} reasoning strategy, avoiding pure enumeration, is thus our key to understanding RLVR generalization. Our intervention, therefore, is not to propose training tasks that are \emph{closer} to the evaluation or simply \emph{harder} than the existing training tasks, but a set of new tasks that more directly incentivize the derivation strategy. In fact, this is also why RLVR could bring improvements to multiple reasoning domains for LLMs in previous works~\citep{liu2025spiral,yuan2025fromfxgx}. Specifically, we propose using bijective functions as $f$ (both injective and surjective) to eliminate uncertainty in non-bijective tasks and encourage derivation via larger certainty.

\mypar{Enhancing OOD Generalization.} Concretely, we co-train on bijective inversion with a $1{:}1$ with the original non-bijective tasks, so this model is called \textbf{\textsc{Mixed}}. As shown in Fig.~\ref{fig:nbj_strategy}, this almost doubles the utilization of derivation strategy compared with the \textsc{Mono} model, and leads to significantly better OOD evaluation (Fig.~\ref{fig:nbj_with_bj}), especially on the novel functions of LenChange. Therefore, the incorporation of bijective inversion effectively improves the RLVR reasoning generalization. We have controlled the data recipe and training configuration for a fair comparison between the two models (details Sec.~\ref{sec:appendix_training_details}) and additionally verified the main conclusions with multiple runs in Sec.~\ref{sec:appendix_seeds}.

\mypar{Why Unexpected?} This is a surprising observation because it contradicts three natural intuitions: (1) bijective tasks ask for a single answer, so they are \emph{further} from the multi-answer evaluation; (2) they are \emph{not clearly harder} by the training reward or by intrinsic difficulty (details in Sec.~\ref{sec:appendix_confounding_factors}); and (3) they do \emph{not} change the in-distribution accuracy notably (Fig.~\ref{fig:main_obs}). These further suggest that the meta-level reasoning strategies are the essential element for RLVR generalization.

For scientific rigor, we have conducted additional experiments in Sec.~\ref{sec:appendix_confounding_factors} to eliminate the confounding factors other than reasoning strategies. Specifically, our controllable experiments show that the improvement is not primarily explained by: (1) a special set of evaluation; (2) increasing the diversity of training tasks; and (3) increasing the difficulty of training tasks. Such analysis further motivates us to \emph{understand} how the reasoning strategies that generalize better are incentivized.

\begin{takeaway}{2}
Incorporating tasks that are even further from the OOD evaluation, \emph{i.e.}, bijective inversion, enhances generalization via encouraging the more effective reasoning strategy.
\end{takeaway}

\section{Bits Per Token Shapes Reasoning Strategies and Generalization}
\label{sec:theory}

With Sec.~\ref{sec:main_observation} revealing reasoning strategies as the central driver of RLVR generalization, this section answers two questions: \textbf{(1)} \emph{what} makes a reasoning strategy effective, \emph{e.g.}, derivation over enumeration, which we frame as \emph{bits of information generated per token} (Sec.~\ref{sec:theory-bits}); and \textbf{(2)} \emph{how} an outcome-only RLVR reward implicitly supervises the reasoning \emph{process} and selects effective strategies (Sec.~\ref{sec:theory-rlvr}). Sec.~\ref{sec:theory-verification} then verifies this mechanism with two interventions.

\subsection{Reasoning Strategies Viewed From Bits per Token}
\label{sec:theory-bits}

\mypar{The Search MDP.} We view a reasoning problem as searching the solution space $\calY$ for a small subset $\ystar \subseteq \calY$ of correct solutions. A strategy, \emph{e.g.}, derivation in our inversion task or reliable calculation in math, is effective to the extent that each emitted token moves towards $\ystar$. We cast this as an MDP: a state $s\in\calS$ is the prompt plus the tokens emitted so far; an action $a\in\calA$ is the next token; transitions $s_{t+1}=f(s_t,a_t)$ are deterministic; trajectories $\tau=(s_0,a_0,\ldots,s_T)$ have bounded length $T$ and terminate at final solutions $y_\tau\in\calY$. The reward is outcome-based, $R(\tau)=\1[y_\tau\in\ystar]$. Identifying a correct solution from $\calY$ thus carries $D := \log(|\calY|/|\ystar|)$ bits of information.

\mypar{Reasoning Strategy and Generalization.} Strategy effectiveness thus becomes these $D$ bits. If a policy creates at most $\log b$ bits per action (\emph{e.g.}, choosing one number from $b$ candidates), we would expect $T\log b \geq D$. Effective reasoning is therefore a higher \emph{bit-rate} towards $\ystar$. \emph{Derivation} contributes more bits per step by pruning candidates structurally, so it generalizes better to harder problems that require more bits compared with \emph{Enumeration} rejecting one candidate at a time.

\subsection{Outcome-based Reward Incentivizes Reasoning Process via Bit Rates}
\label{sec:theory-rlvr}

The bit-rate view characterizes \emph{which} strategies are effective. We now show how outcome-only RLVR \emph{incentivizes} them, first with an actor-critic (Proposition~\ref{prop:pg}) and then with GRPO (Proposition~\ref{prop:grpo}).

\mypar{Connection with RL Advantage.} For a policy $\pi$, the success probability $p^\pi(s):=\Prob_{\tau\sim\pi}[y_\tau\in\ystar\mid s_t=s]$ equals the value function, $V^\pi(s)=p^\pi(s)$, and $p^\pi(s_0)$ is the prompt's success rate. The change $\Delta p_t := p^\pi(s_{t+1})-p^\pi(s_t)$ records the progress of one token: intuitively, how many candidates it prunes (Fig.~\ref{fig:teaser}\textcolor{iccvblue}{b}). Information-theoretically, $-\log p^\pi(s)$ counts the bits still owed at $s$, so one token yields $\log\!\big(p^\pi(s_{t+1})/p^\pi(s_t)\big)$ bits, of which $\Delta p_t$ is the first-order linearization. As the reward is success itself, $\Delta p_t$ also equals the step-level advantage, $A^\pi(s_t,a_t)=Q^\pi(s_t,a_t)-V^\pi(s_t)=\Delta p_t$ (Proposition~\ref{prop:pg}). Thus, a strategy with more information per token, \emph{i.e.}, larger $\Delta p_t$, is credited more. 

\begin{proposition}[Actor-critic]\label{prop:pg}
Under terminal reward and deterministic transitions as Sec.~\ref{sec:theory-bits}, the advantage at every step equals the success increment, $A^\pi(s_t,a_t)=\Delta p_t$, and therefore
\[
\nabla_\theta J(\pi_\theta) \;=\; \E_{\tau}\!\left[\sum_{t=0}^{T-1}\Delta p_t\,\nabla_\theta\log\pi_\theta(a_t\mid s_t)\right].
\]
\end{proposition}
\noindent The proof (Sec.~\ref{sec:appendix_theory_pg}) combines $R(\tau)-p^\pi(s_0)=\sum_t\Delta p_t$ with the value baseline $V^\pi(s_t)=p^\pi(s_t)$. Policy gradients thus credit each token by the bits it produces.

\begin{proposition}[GRPO]\label{prop:grpo}
GRPO sees only group-level rewards $R^{(i)}=R(\tau^{(i)})$ for $G$ rollouts and assigns the trajectory-level advantage $\hat A^{(i)}=(R^{(i)}-\bar R)/\hat\sigma_R$ to every token, giving the gradient estimator $\hat g_{\text{GRPO}} := \tfrac{1}{G}\sum_{i=1}^G \hat A^{(i)}\sum_{t=0}^{T-1}\nabla_\theta\log\pi_\theta(a_t^{(i)}\mid s_t^{(i)})$. With large group $G$,
\[
\E[\hat g_{\text{GRPO}}] \;\approx\; \frac{1}{\sigma(s_0)}\,\nabla_\theta J(\pi_\theta),
\qquad \text{where} \quad\sigma(s_0):=\sqrt{p^\pi(s_0)(1-p^\pi(s_0))}.
\]
\end{proposition}
\noindent This matters because GRPO has no critic, so the trajectory-level $\hat A^{(i)}$ is its only signal. The proof (Sec.~\ref{sec:appendix_theory_grpo}) treats $\bar R$ as a Monte Carlo estimator of $p^\pi(s_0)$ and, by conditioning, recovers the same $\sum_t\Delta p_t\nabla\log\pi$ as Proposition~\ref{prop:pg} up to a rescaling $1/\sigma(s_0)$. Step-level credit by bits per token therefore emerges from group statistics alone.

This explains why bijective functions incentivize derivation better (Sec.~\ref{sec:finding2}): $|\ystar|=1$ sharpens the per-step advantage gap between high- and low-bit-rate strategies. Compared with non-bijective tasks, bijective tasks thus give up uncertainty but separate derivation from enumeration more clearly.

\subsection{Verification Experiments}
\label{sec:theory-verification}

\begin{figure}[h]
  \centering\vspace{-2mm}
  \includegraphics[width=\linewidth]{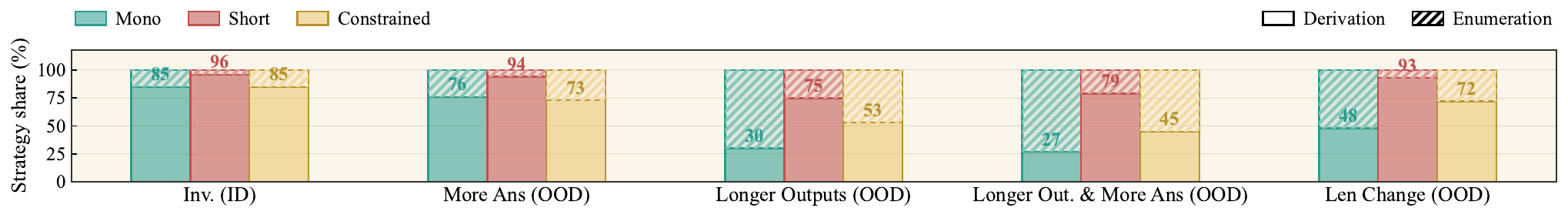}\vspace{-3mm}
  \caption{Tightening bit rates incentivizes the more generalizable derivation strategy. Notation follows Fig.~\ref{fig:nbj_strategy}; ``short'' and ``constrained'' are the shorter $T$ and constrained solution space in Sec.~\ref{sec:theory-verification}.}\vspace{-4mm}
  \label{fig:strategy-budgets}
\end{figure}

The bit-rate view $D/(T\log b)$ predicts that two more interventions should enhance derivation. \textbf{(1)} Shorter $T$: we cut the rollout budget from $8\mathrm{k}$ to $5\mathrm{k}$ tokens to force a higher bit-rate per step. \textbf{(2)} Constrained solution space: we add digit-level constraints to each prompt (\emph{e.g.}, ``the 1st digit cannot be 1, 3, 4, 7''), which also increases $D$. All other settings match \textsc{Mono} in Sec.~\ref{sec:finding1}, and the short budget changes only the maximum response length (Sec.~\ref{sec:appendix_training_details}). Note that both interventions make the problems harder, while our bijective tasks do not necessarily do so. Fig.~\ref{fig:strategy-budgets} shows that both runs raise the derivation rate, validating the link between bit-rate task structure and the strategies RLVR incentivizes (Sec.~\ref{sec:theory-rlvr}). Full results and accuracy curves are in Sec.~\ref{sec:appendix_theory_intervention}. 
\begin{takeaway}{3}
Bits produced per token distinguish effective reasoning strategies for OOD generalization, and also power outcome-only RLVR's incentivization of the reasoning process. 
\end{takeaway}

\section{Behavioral Analysis}

This section conducts a broader inspection of our previous insights into reasoning strategies and OOD generalization. The analyses highlight how the generalization and reasoning strategies could be extended to understand more properties of LLMs.
\subsection{Representation Analysis for Strategy Commitment}
\label{sec:probing}

The first question is whether the strategy shift in Sec.~\ref{sec:main_observation} is visible inside the model's hidden states. Following the probing perspective of \citet{ye2024physics}, we ask \emph{when} the hidden states could reveal the final reasoning strategies reflected in reasoning traces, \emph{i.e.}, derivation or enumeration.

\mypar{Probe setup.} For both \textsc{Mono} and \textsc{Mixed} models, we extract hidden states along each generated chain-of-thoughts on OOD sets and train a linear classifier to predict the final strategy label annotated as Fig.~\ref{fig:appendix_judge_prompt}. In addition, we find that the traces share a stable response structure: restating the problem, reading the function, and then executing a function (see example and details in Sec.~\ref{sec:appendix_probing}). We therefore probe the $N$-th token after $\texttt{<think>}$ for $N\in\{0,100,200,300,500,1000,1500\}$. To avoid shortcuts that only memorize function names or evaluation sets, we report balanced accuracy and compare against a strong null condition and utilizes the \textbf{p-values} to indicate whether the tokens have embedded strategies significantly. Notably, the null shuffles strategy labels only within each function $\times$ evaluation-set subset, so that the shortcut of both functions and evaluatation sets are ruled out (details in Sec.~\ref{sec:appendix_probing}) for statistical rigor.

\begin{figure}[h]
  \centering
  \vspace{-3mm}
  \includegraphics[width=\linewidth]{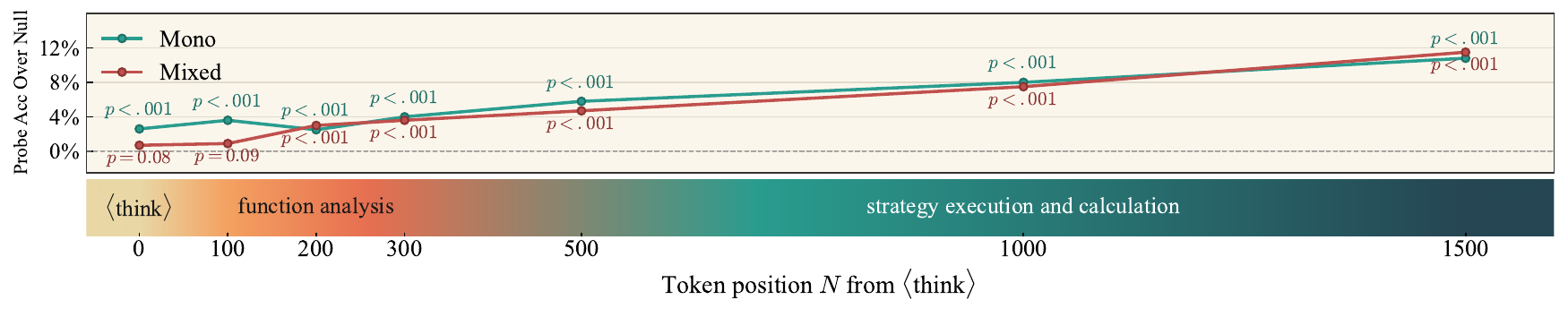}
  \vspace{-6mm}
  \caption{Probing reasoning strategies from hidden states. The y-axis is the delta over a strong null condition (\emph{i.e.}, 76\% for the \textsc{Mixed} model), and we use p-value to read the significance of probing.}
  \label{fig:probe_strategy}
  \vspace{-3mm}
\end{figure}

\mypar{Result.} Fig.~\ref{fig:probe_strategy} shows two interesting patterns. (1) \textsc{Mono}'s strategy is already significantly decodable at $N{=}0$ ($p<.001$), before the generated reasoning has analyzed the function. This matches the collapse in Sec.~\ref{sec:finding1}: the model might enter OOD problems with a preselected behavior rather than a problem-conditioned strategy. (2) \textsc{Mixed} becomes significantly decodable only after the function-analysis region. This suggests that \textsc{Mixed} might commit to a strategy later than \textsc{Mono} and use more task contexts, which is consistent with the observation that \textsc{Mixed}'s reasoning strategy is more robust to OOD shifts. After the function analysis, the two models become similarly decodable, as their traces start to execute the chosen strategy.

\subsection{Mixture of Tasks}
\label{sec:varying_mixture}

The second question is how much bijective data is useful. This ablation tests both sides of the main paper: Sec.~\ref{sec:theory} predicts that bijective tasks should favor higher-bit-rate derivation, while Sec.~\ref{sec:main_observation} leaves open whether more derivation always means better accuracy. We therefore train additional mixtures with bijective:non-bijective ratios $1{:}3$ and $3{:}1$, alongside the original $1{:}1$ \textsc{Mixed} model.

\begin{figure}[tb]
  \centering
  \vspace{-6mm}
  \begin{subfigure}[b]{\linewidth}
    \centering
    \includegraphics[width=\linewidth]{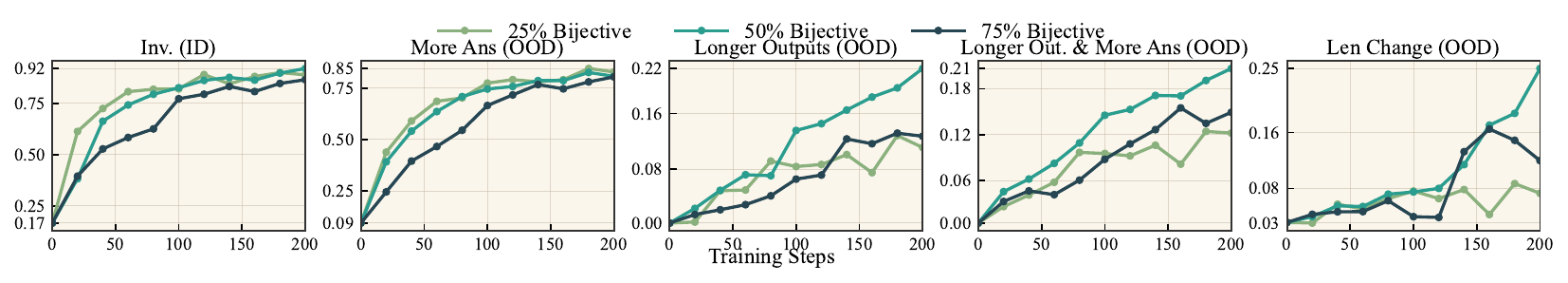}
    \vspace{-6mm}
    \caption{Validation Accuracy}
    \label{fig:mix_ratio}
  \end{subfigure}\\[-2pt]
  \begin{subfigure}[b]{\linewidth}
    \centering
    \includegraphics[width=\linewidth]{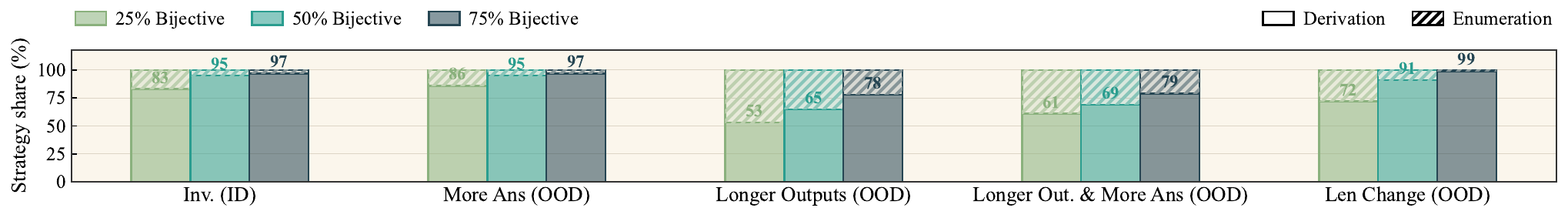}\vspace{-3mm}
    \caption{Reasoning Strategy Distribution}\vspace{-2mm}
    \label{fig:mix_strategy}
  \end{subfigure}
  \caption{Analyzing the recipe of mixtures. (Details in Sec.~\ref{sec:varying_mixture})}\vspace{-2mm}
  \label{fig:mix_ratio_strategy}
\end{figure}

\mypar{Strategy Distribution.} Fig.~\ref{fig:mix_strategy} shows a monotonic effect: increasing the bijective fraction increases the derivation rate. This is the behavior predicted by Sec.~\ref{sec:theory}: RLVR assigns more credit to tokens that structurally shrink the solution space, and this is incentivized by the bijective tasks. The mixture ratio therefore acts as a continuous control over the model's reasoning strategies.

\mypar{Accuracy boundary.} Fig.~\ref{fig:mix_ratio} adds the necessary caveat: the $3{:}1$ mixture uses derivation most often, but the $1{:}1$ mixture is most accurate. Auditing the errors (Sec.~\ref{sec:appendix_mix_errors}) shows that, despite bijective data leading the model to the correct meta-level strategy, the models are prone to execution errors, such as invalid modular arithmetic. The reason is that bijective tasks ask for only one solution, so more bijective data leaves less practice on producing several distinct solutions. This provides additional context for LLM generalization beyond the reasoning-strategy aspect in RLVR generalization: a generalizing model also needs certain task-specific basic skills that make the selected strategy reliable.

\subsection{Analysis of Zero-shot Transfer to Math}
\label{sec:math_zeroshot}

Synthetic RLVR studies sometimes use math benchmarks to hint generalization. However, our goal is not to claim that our synthetic tasks are effective for math, but to test whether the factors identified above, the meta-level strategies and execution skills, also explain behaviors for real-world questions.

\begin{wraptable}{r}{0.5\textwidth}
  \vspace{-4mm}
  \caption{Zero-shot transfer to math benchmark: MATH500~\citep{hendrycks2021math}, Minerva-Math~\citep{lewkowycz2022minerva}, and OlympiadBench~\citep{he2024olympiadbench}. Accuracy in \%.}\vspace{-3mm}
  \label{tab:math_zeroshot}
  \centering
  \small
  \setlength{\tabcolsep}{3pt}
  \begin{tabular}{l@{\hspace{6pt}}|r@{\hspace{4pt}}|r@{\hspace{4pt}}r}
    \toprule
    Benchmark                                        & Qwen          & \textsc{Mono} & \textsc{Mixed} \\
    \midrule
    MATH500 & 76.0          & \textbf{77.2} & 75.8           \\
    Minerva-Math         & 36.0          & 34.9          & \textbf{37.5}  \\
    AIME24                                           & 10.0          & 10.0          & \textbf{13.3}  \\
    AIME25                                           & 3.3           & 0.0           & \textbf{10.0}  \\
    AMC23                                            & 47.5          & 55.0          & \textbf{62.5}  \\
    OlympiadBench         & 37.6          & 38.1          & \textbf{39.9}  \\
    \midrule
    \textbf{Average}                                 & 35.1          & 35.9          & \textbf{39.8}  \\
    \bottomrule
  \end{tabular}
  \vspace{-4mm}
\end{wraptable}
Table~\ref{tab:math_zeroshot} reports zero-shot results (details in Sec.~\ref{sec:appendix_math_protocol}). Both RLVR models improve over the base model on average, but they specialize differently. \textsc{Mono} is stronger on MATH500,better at sustained mechanical execution. \textsc{Mixed} is stronger on the other five harder benchmarks and shows better structural reasoning.

\mypar{Trace Analysis.} We audit the disagreement of the models. \textsc{Mixed} more often preserves the problem structure better, for example, keeping a full divisor set, respecting ordered constraints, or parsing a custom operator correctly. This coincides with the problem-conditioned strategy commitment in Sec.~\ref{sec:probing}. \textsc{Mono}, by contrast, is more reliable when the bottleneck is long mechanical execution, such as polynomial division or checking alternate cases. Secs.~\ref{sec:appendix_math_structure} and~\ref{sec:appendix_math_execution} provides concrete examples. These audits connect the math benchmark numbers back to our central focus on reasoning strategies: \textsc{Mixed} transfers a better meta-level reasoning strategy for the task structures, while \textsc{Mono} retains stronger execution. This suggests why synthetic RL can improve math in prior work~\citep{yuan2025fromfxgx,liu2025spiral}: it transfers a meta-level preference for structural reasoning, while domain-specific execution still needs in-domain training.

\subsection{Agentic Transfer}
\label{sec:agentic}

\begin{wrapfigure}{r}{0.46\textwidth}
  \vspace{-6mm}
  \centering
  \includegraphics[width=\linewidth]{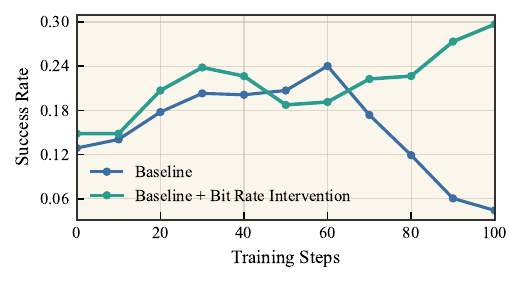}\vspace{-3mm}
  \caption{Our intervention of adding tasks with larger bits rate (Sec.~\ref{sec:main_observation} and Sec.~\ref{sec:theory}) can also help multi-step scenarios.}
  \label{fig:sokoban_dapo}
  \vspace{-4mm}
\end{wrapfigure}

Finally, we ask whether the same principle we used to incentivize higher bit-rate strategies can understand scenarios outside token-level inversion traces. We use Sokoban as a controllable multi-step agentic environment, following RAGEN~\citep{wang2025ragen}, and train with DAPO~\citep{dapo} (details in Sec.~\ref{sec:appendix_agentic}). To create an analogy of the high bit-rate bijective tasks in Sec.~\ref{sec:finding2}, we add a Sokoban variant that succeeds only if the rollout reaches the goal within \emph{optimal\_steps + 2} actions. We mix this variant with the original environment at $1{:}1$. To avoid turning the auxiliary task into a simple difficulty increase, we reduce its grid size by one.

Fig.~\ref{fig:sokoban_dapo} shows that the baseline run degrades, similar to the instability observed in RAGEN~\citep{wang2025ragen}. The mixed run with the auxiliary tasks, however, trains better. Even so, we acknowledge that the multi-step scenario is not a direct analogy with our investigation, as there is no clear categorization of derivation and enumeration in this setting. However, this experiment hints at potential future work: the principle of reasoning strategies and bits per token, though discussed primarily for reasoning tasks in this paper, may also improve RLVR in agentic settings.

\section{Conclusion}
\label{sec:conclusion}

This paper investigates reasoning strategies as a crucial factor for RLVR generalization. On a controllable framework of non-bijective string equations with clearly defined OOD generalization, RLVR teaches the derivation strategy in-distribution, yet the model falls back to ineffective enumeration on OOD tasks. Co-training with bijective tasks restores derivation and enhances OOD generalization, which motivates our theory: effective reasoning strategies contribute more bits per token and are implicitly rewarded by RLVR. Our broader analyses further support the role of reasoning strategies in LLM generalization, and we discuss limitations in Sec.~\ref{sec:appendix_limitations}. We hope our study can inspire RL recipes that improve generalization by controlling reasoning strategies.

\subsection*{AI Use Statement}
We use Gemini-3.0-Flash as an LLM judge to label reasoning strategies (Sec.~\ref{sec:finding1}), and validate it against manual annotation (Sec.~\ref{sec:appendix_strategy_categorization}). We use generative AI tools for editing the writing, \emph{e.g.}, grammar, spelling, and word choice. We have reviewed all AI-assisted content and take responsibility for the final content of this work.

\subsection*{Reproducibility Statement}
Sec.~\ref{sec:appendix_prompt_scoring} gives the prompt format and the verifier, Sec.~\ref{sec:appendix_function_lists} lists every function used for training and evaluation, and Sec.~\ref{sec:appendix_training_details} lists all RL hyperparameters and hardware. The strategy judge prompt is in Fig.~\ref{fig:appendix_judge_prompt}, and the probing protocol is in Sec.~\ref{sec:appendix_probing}. Sec.~\ref{sec:appendix_seeds} reports repeated runs with different seeds and data.

{
\small
\bibliographystyle{iclr2027_conference}
\bibliography{references}
}

\newpage
\appendix
\makeatletter
\@addtoreset{figure}{section}
\@addtoreset{table}{section}
\@addtoreset{equation}{section}
\makeatother
\renewcommand{\thefigure}{\thesection.\arabic{figure}}
\renewcommand{\thetable}{\thesection.\arabic{table}}
\renewcommand{\theequation}{\thesection.\arabic{equation}}

\section{Details of Research Framework}
\label{sec:appendix_research_framework}

\subsection{Prompt Format and Scoring}
\label{sec:appendix_prompt_scoring}

All main experiments use a shared format that is similar to \citet{yuan2025fromfxgx}, with the modifications to inversion prompts. Each prompt contains the full source code for a function, a target output string $y$, and an instruction to return $k$ different digit-string inputs whose execution produces $y$. The model must rely on its own calculation and analysis. Fig.~\ref{fig:appendix_prompt_example} shows a representative prompt drawn from the length-changing OOD evaluation set.

\begin{figure}[ht]
\centering
\begin{tcblisting}{
  enhanced, breakable,
  colback=gray!5, colframe=gray!55,
  arc=2mm, boxrule=0.5pt,
  left=8pt, right=8pt, top=4pt, bottom=4pt,
  listing only,
  listing options={
    basicstyle=\ttfamily\scriptsize,
    breaklines=true,
    columns=fullflexible,
    keepspaces=true,
    showstringspaces=false,
  },
}
You are given a code:

from math import gcd

def func(s: str) -> str:
    """Reduce length by 1: first+last mod 5 becomes new first; drop last.

    Example structure (requested): mix first/last, then remove last digit.
      y[0]   = (x[0] + x[n-1]) % 5
      y[1:]  = x[1:n-1] % 5   (per-digit mod 5 view)
    """
    n = len(s)
    if n <= 1:
        return ""
    x0 = ord(s[0]) - ord("0")
    xl = ord(s[-1]) - ord("0")
    out = [str((x0 + xl) % 5)]
    for ch in s[1:-1]:
        out.append(str((ord(ch) - ord("0")) % 5))
    return "".join(out)


def main_solution(x):
    return func(x)

You are told that `main_solution(x)` returns the string "424331344".
Find 3 DIFFERENT digit-string inputs `x` such that `main_solution(x)` returns exactly that string.

Please reason between the XML tags of <think> </think>, and put the answers, which are ONLY the digit strings, between <answer> </answer> tags separated by commas.
The final response should look like <think> ... </think><answer> x1,x2,... </answer>
\end{tcblisting}
\vspace{-6pt}
\caption{Example prompt for training and evaluation.}
\label{fig:appendix_prompt_example}
\end{figure}

The answer is parsed from \texttt{<answer>} tags as a comma-separated list of digit strings. If fewer than $k$ valid preimages exist, the prompt instructs the model to output only the valid ones. The verifier executes the reference implementation on each submitted candidate rather than comparing against a fixed answer set. This is necessary for non-bijective functions, where the generator may know several valid preimages but need not enumerate all of them. In our evaluation function, we also explicitly conduct the de-duplication of the solutions from the LLMs, to avoid any form of reward hacking.

\subsection{Function Lists}
\label{sec:appendix_function_lists}

We use three function families.
First, the non-bijective functions are the core inversion tasks introduced in Sec.~\ref{sec:task_design} and used to train \textsc{Mono} in Sec.~\ref{sec:finding1}. For convenience, we use the canonical names \texttt{nb\_00}--\texttt{nb\_23}, with 18 training functions and 6 held-out functions.
Second, the bijective functions are the stricter auxiliary tasks introduced in Sec.~\ref{sec:finding2}. They use the names \texttt{func\_00}--\texttt{func\_23}, with 18 training functions and 6 held-out functions.
Third, the length-changing OOD functions are introduced as the \textsc{Length Change} evaluation axis in Sec.~\ref{sec:task_design}. They use the names \texttt{lenop\_00}--\texttt{lenop\_05}.
We intentionally make the non-bijective held-out split easier to further support our observed OOD reasoning strategy collapse: the training and held-out functions share primitives such as cyclic neighbor operations, prefix/state updates, and modular or folded emissions. In the prompts, the displayed function is wrapped as \texttt{func}; the names below are dataset identifiers.

\mypar{Non-bijective train functions.}
\texttt{nb\_00}: cyclic floor-average of neighbors.
\texttt{nb\_01}: cyclic ceil-average of neighbors.
\texttt{nb\_03}: cyclic capped sum.
\texttt{nb\_04}: cyclic overflow amount past 9.
\texttt{nb\_05}: cyclic distance of neighbor-sum to 9.
\texttt{nb\_07}: cyclic folded neighbor-difference mod 10.
\texttt{nb\_08}: cyclic average plus absolute difference mod 10.
\texttt{nb\_10}: cyclic triple average floor.
\texttt{nb\_11}: cyclic triple average ceil.
\texttt{nb\_13}: cyclic folded triple sum mod 10.
\texttt{nb\_15}: cyclic clamped absolute value of $a+b-c$.
\texttt{nb\_16}: prefix sum to evenized emission.
\texttt{nb\_17}: prefix sum to folded emission.
\texttt{nb\_19}: affine state update plus folded emission.
\texttt{nb\_20}: cyclic median of three.
\texttt{nb\_21}: cyclic rank of the first element in a triple.
\texttt{nb\_22}: cyclic capped neighbor product.
\texttt{nb\_23}: cyclic local total variation.

\mypar{Non-bijective held-out functions.}
\texttt{nb\_02}: cyclic absolute difference.
\texttt{nb\_06}: cyclic folded neighbor-sum mod 10.
\texttt{nb\_09}: cyclic two-view merge.
\texttt{nb\_12}: cyclic triple range.
\texttt{nb\_14}: cyclic second-difference magnitude mod 10.
\texttt{nb\_18}: prefix sum to square emission.

\mypar{Bijective train functions.}
\texttt{func\_00}: reverse, then add 1 to every digit.
\texttt{func\_01}: rotate left by 1, then add 3 to every digit.
\texttt{func\_02}: swap first and last characters, then add 3.
\texttt{func\_03}: swap adjacent pairs, then add 1.
\texttt{func\_04}: swap the two halves, then add 2.
\texttt{func\_05}: reverse each half, then add 3.
\texttt{func\_06}: move odd-indexed characters to the front, then add 1.
\texttt{func\_07}: interleave the two halves, then add 2.
\texttt{func\_08}: reverse the inner substring, then add 3.
\texttt{func\_09}: rotate each full group of three right by 1, then add 1.
\texttt{func\_10}: reverse, then add 1 to even positions and 2 to odd positions.
\texttt{func\_11}: rotate left by 1, then add 1 or 3 on alternating positions.
\texttt{func\_12}: add $i+1$ to the digit at position $i$.
\texttt{func\_13}: add $2i+1$ to the digit at position $i$.
\texttt{func\_14}: add 1 on the left half, 3 on the right half, and 2 in the middle.
\texttt{func\_15}: add the repeating pattern 1,2,3.
\texttt{func\_16}: reverse, then add $i+1$ at position $i$.
\texttt{func\_17}: swap adjacent pairs, then add $i+2$ at position $i$.

\mypar{Bijective held-out functions.}
\texttt{func\_18}: move odd-indexed characters to the front, then apply half-specific shifts.
\texttt{func\_19}: swap halves, then add the repeating pattern 1,2,3.
\texttt{func\_20}: running sum with $+1$ drift.
\texttt{func\_21}: running sum with $+2$ drift.
\texttt{func\_22}: running sum with alternating $+1/+2$ drift.
\texttt{func\_23}: right-to-left running sum with $+1$ drift.

\mypar{Length-changing OOD functions.}
\texttt{lenop\_00}: reduce length by one via non-cyclic neighbor-sum mod 5.
\texttt{lenop\_01}: reduce length by one via non-cyclic signed neighbor mix mod 5.
\texttt{lenop\_02}: reduce length by one by emitting first-plus-last mod 5, then dropping the last digit.
\texttt{lenop\_03}: increase length by one via cyclic neighbor-sum mod 5 plus a checksum digit.
\texttt{lenop\_04}: increase length by one via a mod-5 affine state machine plus a final-state digit.
\texttt{lenop\_05}: increase length by one via mod-5 prefix sums plus a start/end consistency check.

\subsection{Training Details}
\label{sec:appendix_training_details}

\mypar{Implementation.}
We run RL experiments with VERL/HybridFlow~\citep{sheng2024hybridflow}, using the GRPO objective~\citep{shao2024deepseekmath} with the DAPO-style filtering of zero-reward-variance groups~\citep{dapo}.
The policy is initialized from Qwen2.5-7B-Instruct; rollouts are generated with \texttt{vllm}, and actor/reference models are sharded with FSDP.
Training is conducted on one node with 8 NVIDIA H100 80GB GPUs, approximately 1 day.

\mypar{Reward.} We follow common RLVR practice and construct the reward by 0.1 of format reward, including the thinking format and number of solutions, and 0.9 of the accuracy scoring.

\mypar{Core hyperparameters.}
Unless otherwise stated, we train for 200 update steps with a prompt batch size of 16, one PPO epoch per update, actor learning rate $10^{-6}$, clip ratio $0.2$, entropy coefficient $10^{-3}$, and gradient clipping at $1.0$.
We use GRPO groups of 16 sampled rollouts per prompt, with sampling temperature $1.0$, top-$p$ of $1.0$, and no top-$k$ truncation.
The maximum prompt length is 1{,}024 tokens and the maximum response length is 8{,}192 tokens; the short-budget intervention in Sec.~\ref{sec:theory-verification} changes only this response budget to $5\mathrm{k}$ tokens.
We do not use a learned reward model or KL penalty in these controlled experiments; rewards are computed by the verifier in Sec.~\ref{sec:appendix_prompt_scoring}.
Validation is run every 20 steps with greedy decoding (\texttt{temperature}=0, one rollout per prompt).

\subsection{Selection of Base Model}
\label{sec:appendix_base_models}

This section explains why we decided to use Qwen2.5-7B-Instruct as our base model. For controlled studies of LLM post-training, the base model is part of the experimental design: if it is too weak, failures mostly measure missing task competence; if it is too strong, the task saturates before RL exposes what the reward is selecting.
This is the same methodological motivation behind prior controlled reasoning studies~\citep{ye2024physics,yuan2025fromfxgx,wang2025ragen}, which choose settings where the target behavior is observable, the verifier is clean, and the training dynamics remain diagnostic. Since our goal is to reveal a nuanced RLVR generalization failure rather than to maximize absolute accuracy, the appropriate base model is one that can solve the in-distribution task after RL but still leaves enough room for the learned reasoning strategy to vary and collapse OOD.

\mypar{Additional Base Models.}
We tried Llama-3.1-8B-Instruct~\citep{grattafiori2024llama} and Qwen3-4B-Instruct~\citep{yang2025qwen3} under the same inversion task setup. Llama-3.1-8B-Instruct does not provide a suitable initialization for our question: after RL on non-bijective tasks, it reaches only $27.1\%$ on in-distribution non-bijective inversion. This performance indicates that the model falls into the category of less capable base model, has not yet acquired reliable inversion reasoning or the basic execution skills needed for our analysis. Failures in this regime would mostly reflect insufficient task competence, rather than a strategy-level OOD generalization gap. So we exclude Llama-3.1-8B-Instruct as our base model.

Qwen3-4B-Instruct has the opposite issue: it is already too strong for this controlled study. Its in-distribution non-bijective inversion accuracy reaches $97.8\%$ after just 20 steps of RL training, indicating that our controlled study has not significantly influenced the model's behavior yet. This leaves little room for RL to reveal a meaningful strategy-acquisition process, because the model is close to saturation before the training dynamics become informative.

\mypar{Conclusion.} Using controlled study to investigate LLM is a co-design across the model, task, data.
For our purpose, Qwen2.5-7B-Instruct is the appropriate operating point because it makes the central comparison identifiable: after RL, both \textsc{Mono} and \textsc{Mixed} solve the in-distribution non-bijective inversion task at about $91$--$92\%$, so the source-task competence, prompt understanding, and verifier-compatible answer behavior are already established. The remaining failure is therefore not that the model cannot solve the task, but that the reasoning strategy learned on the source distribution does not necessarily transfer OOD. This is exactly the phenomenon our paper aims to reveal. If we used a weaker model, the same OOD failures would be confounded with missing basic skills; if we used a saturated model, RL would provide too little behavioral change to analyze strategy selection. Making the task much longer or harder would similarly mix our target failure with arithmetic, context-length, and execution errors. Thus, Qwen2.5-7B-Instruct is a principled choice for exposing reasoning-strategy collapse under already favorable conditions for RLVR generalization.

\section{Details of Revealing Strategy Collapse and Fixes on OOD Generalization}
\subsection{Reasoning Strategy Categorization}
\label{sec:appendix_strategy_categorization}

\begin{figure}[h]
\centering
\begin{tcblisting}{
  enhanced, breakable,
  colback=gray!5, colframe=gray!55,
  arc=2mm, boxrule=0.5pt,
  left=8pt, right=8pt, top=4pt, bottom=4pt,
  listing only,
  listing options={
    basicstyle=\ttfamily\tiny,
    breaklines=true,
    columns=fullflexible,
    keepspaces=true,
    showstringspaces=false,
  },
}
You are a reasoning-pattern classifier for digit-string inverse-function problems. You will read a model's step-by-step reasoning and classify it into exactly one category.

Categories:

(A) DERIVE -- Structured constraint solving
The model identifies dependency structure: it chooses SOME digit positions freely, then DEDUCES the remaining digits from constraint equations. The key test: are most digit values produced as outputs of equations (derived) rather than chosen independently (guessed)?
- Choosing one free variable then computing the rest from equations IS derive.
- Setting up equations AND using them to solve for unknowns IS derive.

(B) ENUM -- Brute-force trial of full assignments
The model assigns ALL or nearly all digit positions by guessing, then checks if the assignment works. The workflow is: guess a complete input -> verify -> discard or keep -> guess again.
- Guessing complete strings and checking output IS enum.
- Setting up equations but then IGNORING them and guessing complete strings IS enum.
- Trying many different digit-tuple assignments one by one and checking each IS enum.
- Trying values without using prior constraints to narrow choices IS enum.
- For functions like max-min-of-triplets or second-order-difference, if the model guesses complete digit patterns and checks them, that IS enum even if equations were written down first.

(C) LOOP -- Stuck repetition
The model repeats the same or nearly identical computations/attempts multiple times without making progress. LOOP is a degenerate form of enumeration where the model gets stuck.
Key signals:
  1. The SAME digit values or computation blocks appear multiple times.
  2. The model keeps getting the same wrong result but retries similar values.
  3. The response ends abruptly mid-computation after repeated attempts.
  4. The latter reasoning recycles earlier content without finding new solutions. A response can start with valid derive or enum reasoning but degenerate into a loop. If the model spends most of its token budget stuck in repetition, classify as LOOP. A very long response (>10000 chars) with NO <answer> tag and repeated computation blocks is LOOP. A long response that makes genuine progress trying DIFFERENT values is ENUM, not LOOP.

Grading procedure:
1. Check for LOOP. If yes -> C.
2. If not LOOP, determine DERIVE vs ENUM. Find where the model starts proposing solution digit values. If most values are outputs of constraint equations -> A. If most are independently guessed and checked -> B.

Output format: Answer with ONLY the single letter A, B, or C. Do not explain.

USER MESSAGE:
PROBLEM:
{problem}

RESPONSE:
{response}

Classify the reasoning strategy. Answer A, B, or C only.
\end{tcblisting}
\vspace{-4mm}
\caption{LLM prompt for reasoning strategy categorization.}\vspace{-4mm}
\label{fig:appendix_judge_prompt}
\end{figure}

\mypar{Why Derivation and Enumeration.} As discussed in Sec.~\ref{sec:finding1} and illustrated in Fig.~\ref{fig:strategy_examples}, our analysis separates inversion reasoning into \emph{analytical derivation} and \emph{brute-force enumeration}. This categorization is exhaustive for our purpose because the core decision is binary: when proposing the input digits, does the LLM rely on enumerating full digit assignments and then verify them? If most digit values are outputs of equations, we classify the trace as derivation; if most digit values are chosen independently and checked as candidate strings, we classify it as enumeration. Realistic traces can mix derivation, case analysis, local enumeration, and verification. The binary label does not deny such mixtures: a trace that solves most digits from equations, possibly with case analysis over a few free digits, counts as derivation. The label therefore isolates the pattern we study, \emph{i.e.}, whether the model abandons structural reasoning. 

\mypar{Automatic Categorization with LLM Judges.}
We use Gemini-3.0-Flash as an LLM judge to label reasoning traces with a deliberately simple output space, following the SimpleQA~\citep{wei2024simpleqa} style of asking the grader for a single unambiguous label. Internally, we use three labels: DERIVE, ENUM, and LOOP. LOOP has not appeared in the main paper, since it is a degenerate descendant of ENUM and was used primarily for our error analysis. It mostly shows the patterns where the model repeatedly enumerates or checks similar candidates until it gets stuck. For all main-paper strategy plots, LOOP is merged into ENUM because both reflect failure to perform structural derivation. To validate the precision of the judge, we manually annotated 100 randomly sampled traces and found $96\%$ agreement with the LLM judge. This $4\%$ disagreement is much smaller than the derivation gap between \textsc{Mono} and \textsc{Mixed} on Length Change ($48\%$ \emph{vs.}\ $91\%$), so our conclusions are robust to judge errors. We label all generated responses, both correct and incorrect, so the strategy shares are not biased toward successful traces. The prompt is included in Fig.~\ref{fig:appendix_judge_prompt}.

\subsection{Confounding Factors in Fixing Generalization}
\label{sec:appendix_confounding_factors}

In Sec.~\ref{sec:main_observation}, we have shown that the reasoning strategy collapse of the \textsc{Mono} model can be fixed by introducing an auxiliary set of bijective tasks with stricter requirements in terms of bit-rates per token (as in Sec.~\ref{sec:theory}). To validate that our discovery is of scientific rigor and commonness, we discover three common confounding factors that could contribute the phenomenon. In fact, they are not contributing factors to better OOD generalization; thus, this section better supports our observations of reasoning strategy collapse and generalization in RLVR.
\subsubsection{Training Task Difficulty}
\label{sec:appendix_confounding_task_difficulty}
\begin{figure}[!h]
  \centering
  \includegraphics[width=0.62\linewidth]{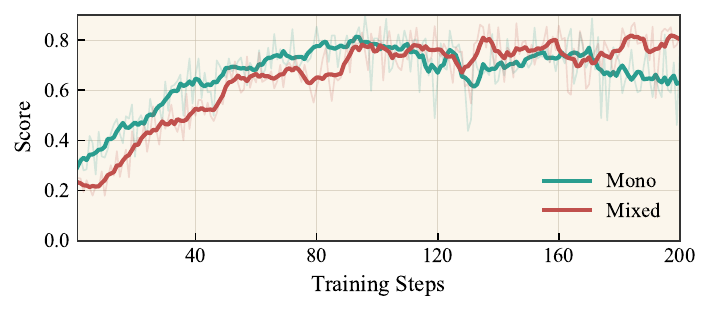}
  \vspace{-2mm}
  \caption{Training-time accuracy for \textsc{Mono} and \textsc{Mixed} models.}
  \label{fig:training_score_confounding}
  \vspace{-4mm}
\end{figure}

A possible alternative explanation is that bijective co-training improves OOD performance by making the RL objective substantially harder, rather than by changing the rewarded reasoning strategy. Fig.~\ref{fig:training_score_confounding} tests this explanation using the training-time verifier score. The curve does show an early optimization lag for \textsc{Mixed}. However, this lag is transient: the reward of the two model becomes similar, and the overall reward of \textsc{Mixed} is even higher eventually . Thus the reward signal does not support a difficulty explanation in which \textsc{Mixed} generalizes better simply because it is trained on a uniformly harder objective. In addition, the added bijective tasks are also further from the non-bijective OOD evaluations, since they ask for a single $x$ rather than multiple valid answers. Therefore, the improvement of \textsc{Mixed} cannot be reduced to either a closer evaluation format or a consistently harder training objective.

\mypar{Intrinsic Difficulty.} Training reward depends on the model, so we also compare the functions directly. The bijective functions (Sec.~\ref{sec:appendix_function_lists}) mainly consist of reversible position permutations, digit-wise modular shifts, and invertible running-sum variants, while the non-bijective functions mainly introduce many-to-one interactions between neighboring digits. Consider the cost of an optimal inversion program. A representative bijective function reverses the input and adds 1 modulo 10 to every digit; its inverse is deterministic and takes $O(n)$ modular operations. A representative non-bijective function emits the cyclic floor-average of neighbors, $y_i=\lfloor(x_i+x_{i+1})/2\rfloor$; each output digit admits multiple predecessors, so inversion branches at ambiguous positions and costs $O(2^k n)$, where $k$ is the number of branching positions. Thus, the bijective functions are not intrinsically harder. For an LLM, difficulty also depends on the adopted strategy: pure enumeration is \emph{harder} on bijective tasks, where only one of $10^n$ inputs is correct. This strategy dependence is why we also report the training reward above.

\subsubsection{Another Set of Functions}
\label{sec:appendix_confounding_another}

We also show similar observations as Sec.~\ref{sec:main_observation} with another set of functions, especially changing the set of non-bijective functions for training and evaluation. Based on the new functions, we re-train  new \textsc{Mono} and \textsc{Mixed} models.

\mypar{Function List.}
This confounding study uses a balanced alternative non-bijective family with the same split size as Sec.~\ref{sec:appendix_function_lists}: 18 training functions and 6 held-out functions for OOD evaluation. Same as before, we use \texttt{nb\_xx} for the convenience of notations. \textbf{Training}:
\texttt{nb\_00}: cyclic floor-average of neighbors, then add 1.
\texttt{nb\_01}: cyclic ceil-average of neighbors, then add 2.
\texttt{nb\_03}: cyclic capped neighbor sum, then add 1.
\texttt{nb\_04}: cyclic floor-average of triples, then add 1.
\texttt{nb\_05}: cyclic ceil-average of triples, then add 2.
\texttt{nb\_07}: cyclic triple median, then add 2.
\texttt{nb\_08}: mirror-pair floor-average, then add 1.
\texttt{nb\_09}: mirror-pair ceil-average, then add 2.
\texttt{nb\_11}: mirror-pair maximum, then add 2.
\texttt{nb\_12}: sort adjacent pairs ascending, then add 1.
\texttt{nb\_13}: sort adjacent pairs descending, then add 2.
\texttt{nb\_15}: sort full triplets descending, then add 2.
\texttt{nb\_16}: prefix minimum, then add 1.
\texttt{nb\_17}: prefix maximum, then add 1.
\texttt{nb\_19}: suffix maximum, then add 2.
\texttt{nb\_20}: prefix-sum state with floor-half emission, then add 3.
\texttt{nb\_21}: prefix-sum state with folded-distance emission, then add 1.
\texttt{nb\_23}: prefix-sum state with distance-to-five emission. 
\textbf{Evaluation:}
\texttt{nb\_02}: cyclic neighbor minimum, then add 1.
\texttt{nb\_06}: cyclic triple minimum, then add 1.
\texttt{nb\_10}: mirror-pair minimum, then add 1.
\texttt{nb\_14}: sort full triplets ascending, then add 1.
\texttt{nb\_18}: suffix minimum, then add 2.
\texttt{nb\_22}: prefix-sum state with squared-state emission, then add 1.

\begin{figure}[t]
  \centering
  \begin{subfigure}[b]{\linewidth}
    \centering
    \includegraphics[width=\linewidth]{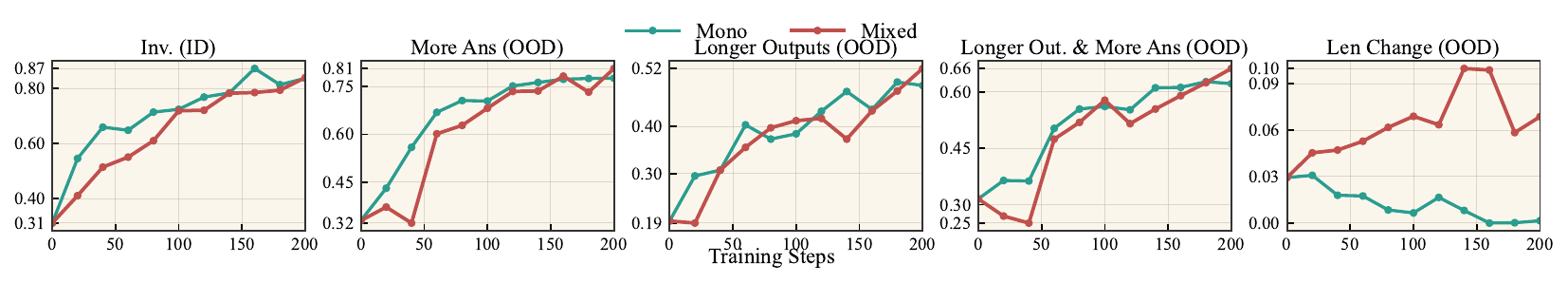}
    \vspace{-6mm}
    \caption{Evaluation Accuracy During RL}
    \label{fig:confounding_another}
  \end{subfigure}\\[-2pt]
  \begin{subfigure}[b]{\linewidth}
    \centering
    \includegraphics[width=\linewidth]{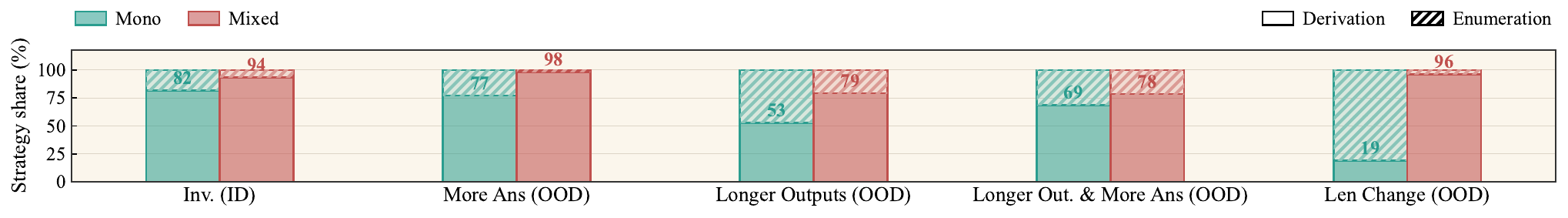}\vspace{-3mm}
    \caption{Reasoning Strategy Distribution}\vspace{-2pt}
    \label{fig:confounding_another_strategy}
  \end{subfigure}
  \caption{Confounding experiments via training on a new set of functions. The distribution of reasoning strategies and better OOD accuracy of \textsc{Mixed} validates our conclusion in Sec.~\ref{sec:main_observation}.}
  \label{fig:confounding_another_combined}
  \vspace{-3mm}
\end{figure}

\mypar{Analysis.} As shown in Fig.~\ref{fig:confounding_another_combined}, the \textsc{Mixed} model is still better than \textsc{Mono} model on OOD evaluations, with similar performance for in-distribution cases. This is consistent with our observations in Sec.~\ref{sec:main_observation}. Notably, the difference in the most challenging LenChange set marks a drastic difference in OOD generalization. In addition, these observations are accompanied with the consistent observations on reasoning strategies, where the \textsc{Mixed} model more robustly utilizes the derivation reasoning strategy, and leads to better OOD performance. Such an observation suggests that our observations in Sec.~\ref{sec:main_observation} is generalizable.

Regarding the performance, we also would like to share that this additional set of non-bijective functions are easier for Qwen2.5-7B model, which leads to much better performance on the OOD evaluation like longer outputs and more answers: in fact, the model suffers less from harder OOD generalization in this case. Such an observation also points to a potential future work as to how to control the difficulty levels of the functions themselves in the controllable studies of RLVR. 

\subsubsection{Training Task Diversity}
\label{sec:appendix_confounding_diversity}

\begin{figure}[t]
  \centering
  \begin{subfigure}[b]{\linewidth}
    \centering
    \includegraphics[width=\linewidth]{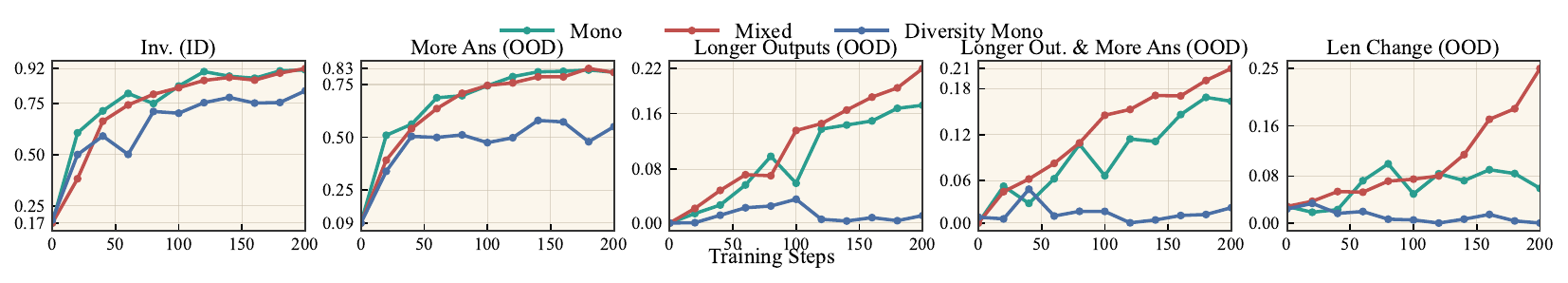}
    \vspace{-6mm}
    \caption{Evaluation Accuracy During RL}
    \label{fig:confounding_diversity}
  \end{subfigure}\\[-2pt]
  \begin{subfigure}[b]{\linewidth}
    \centering
    \includegraphics[width=\linewidth]{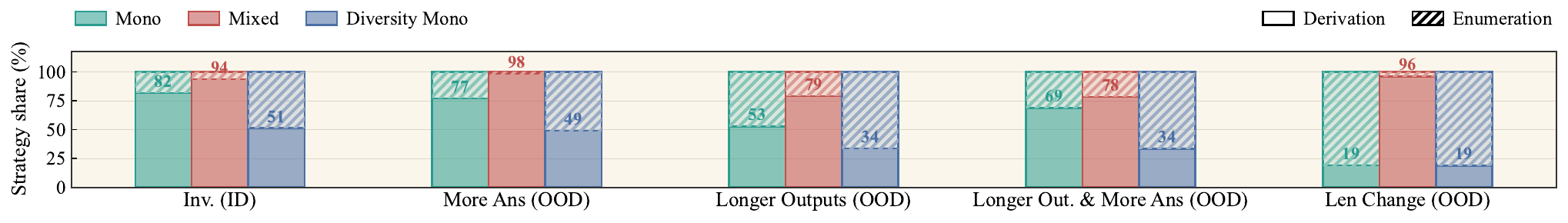}\vspace{-3mm}
    \caption{Reasoning Strategy Distribution}\vspace{-2pt}
    \label{fig:confounding_diversity_strategy}
  \end{subfigure}
  \caption{Confounding experiment via increasing the diversity of non-bijective functions for training. \textsc{Diversity Mono} does not achieve the effect as \textsc{Mixed}, validating our Sec.~\ref{sec:main_observation}.}
  \label{fig:confounding_diversity_combined}
  \vspace{-3mm}
\end{figure}

We finally validate whether \textsc{Mixed} performs better not because of its bijectiveness, but the additional diversity in the training tasks. We therefore adopt the additional functions in Sec.~\ref{sec:appendix_confounding_another} into the original \textsc{Mono} training, so that it possesses the same number of functions in the training tasks as \textsc{Mixed}. As shown in Fig.~\ref{fig:confounding_diversity}, these additional functions do not lead to better performance. Instead, they dilute the learning signals inside the original set of functions, which have shared function primitives. So enlarging the non-bijective function set does not by itself induce derivation, and the OOD gain of \textsc{Mixed} is not explained by greater training-task diversity.

\subsubsection{Repeated Runs}
\label{sec:appendix_seeds}

We repeat the \textsc{Mono} and \textsc{Mixed} runs with two more random seeds and differently sampled training data, keeping all other settings unchanged. Table~\ref{tab:appendix_seeds} reports the final accuracy and derivation share on Length Change, the largest OOD shift. Every run reproduces the main observation: \textsc{Mixed} reaches several times higher accuracy and uses derivation far more often than \textsc{Mono}.

\begin{table}[h]
  \centering
  \small
  \caption{Repeated runs on Length Change. Higher is better for both metrics. ``Main'' is the run in Fig.~\ref{fig:main_obs}.}
  \label{tab:appendix_seeds}
  \begin{tabular}{lcccc}
    \toprule
     & \multicolumn{2}{c}{Accuracy (\%)} & \multicolumn{2}{c}{Derivation (\%)} \\
    \cmidrule(lr){2-3}\cmidrule(lr){4-5}
    Run & \textsc{Mono} & \textsc{Mixed} & \textsc{Mono} & \textsc{Mixed} \\
    \midrule
    Seed 0 & 5.7 & \textbf{21.6} & 46.0 & \textbf{95.1} \\
    Seed 1 & 4.1 & \textbf{18.3} & 50.8 & \textbf{93.2} \\
    Main   & 6.0 & \textbf{25.2} & 48.0 & \textbf{91.6} \\
    \midrule
    Mean $\pm$ std & 5.3 $\pm$ 1.0 & \textbf{21.7 $\pm$ 3.5} & 48.3 $\pm$ 2.4 & \textbf{93.3 $\pm$ 1.8} \\
    \bottomrule
  \end{tabular}
\end{table}

\section{Details of Theoretical Analysis in Sec.~\ref{sec:theory}}
\label{sec:appendix_theory}

This section gives the proof details for Propositions~\ref{prop:pg} and~\ref{prop:grpo}. We use the search-MDP notation from Sec.~\ref{sec:theory}: transitions are deterministic, a trajectory $\tau$ terminates with a candidate solution $y_\tau\in\calY$, and the outcome reward is $R(\tau)=\1[y_\tau\in\ystar]$ for the correct-solution set $\ystar\subseteq\calY$. For any state $s$, the success probability
$p^\pi(s)=\Prob_{\tau\sim\pi}[y_\tau\in\ystar\mid s_t=s]$
is the value function $V^\pi(s)$. The central quantity is the one-token success-probability increment
\[
\Delta p_t := p^\pi(s_{t+1})-p^\pi(s_t),
\]

which is the probability-scale version of the bits-per-token signal in Sec.~\ref{sec:theory-bits}: positive $\Delta p_t$ means that action $a_t$ moves the current reasoning state closer to a correct terminal solution.

\subsection{Proof of Proposition~\ref{prop:pg}: Advantages Are Success Increments in Actor-Critic}
\label{sec:appendix_theory_pg}

\begin{lemma}[Success-increment decomposition]\label{lem:telescope}
For any complete trajectory $\tau$ generated under policy $\pi$,
\[
R(\tau) - p^\pi(s_0) \;=\; \sum_{t=0}^{T-1} \Delta p_t .
\]
\end{lemma}

\begin{proof}
The sum on the right collapses to $p^\pi(s_T)-p^\pi(s_0)$. At a terminal state, the final solution is already fixed, so $p^\pi(s_T)=\1[y_\tau\in\ystar]=R(\tau)$.
\end{proof}

The lemma rewrites the trajectory-level outcome, relative to the prompt baseline, as accumulated per-token progress. Starting from REINFORCE~\citep{williams1992simple} for $J(\pi_\theta)=\E_\tau[R(\tau)]$,
\[
\nabla_\theta J(\pi_\theta) = \E_{\tau}\!\left[R(\tau) \sum_{t=0}^{T-1} \nabla_\theta \log \pi_\theta(a_t \mid s_t)\right],
\]

and using the value baseline $V^\pi(s_t)=p^\pi(s_t)$, the policy-gradient theorem gives
\begin{equation}\label{eq:pg-app}
\nabla_\theta J(\pi_\theta) = \E_{\tau}\!\left[\sum_{t=0}^{T-1} A^\pi(s_t, a_t)\,\nabla_\theta \log \pi_\theta(a_t \mid s_t)\right],\quad A^\pi(s_t,a_t)=Q^\pi(s_t,a_t)-V^\pi(s_t).
\end{equation}

With terminal reward and deterministic transitions, taking $a_t$ fixes the next state, so
$Q^\pi(s_t,a_t)=V^\pi(s_{t+1})=p^\pi(s_{t+1})$.
Therefore $A^\pi(s_t,a_t)=p^\pi(s_{t+1})-p^\pi(s_t)=\Delta p_t$. Substituting this identity into~\eqref{eq:pg-app} proves Proposition~\ref{prop:pg}.\hfill$\square$

\mypar{Implication.}
Proposition~\ref{prop:pg} is the formal bridge between bits per token and RL credit. In our search view, a token receives positive advantage exactly when it increases the probability of eventually reaching a correct solution. A reasoning strategy that reliably prunes the candidate space, such as analytical derivation in our inversion tasks, therefore creates repeated positive increments; brute-force enumeration creates much weaker increments except when a guessed candidate happens to land near a solution. This is the sense in which outcome reward can still favor an internal reasoning process.

\subsection{Proof of Proposition~\ref{prop:grpo}: GRPO}
\label{sec:appendix_theory_grpo}

\mypar{Estimator.}
GRPO does not use a learned critic to assign per-token advantage. For each prompt $s_0$, it samples a group of $G$ independent rollouts $\{\tau^{(i)}\}_{i=1}^G\sim\pi_\theta$, observes rewards $R^{(i)}:=R(\tau^{(i)})$, and computes
\[
\bar R := \frac{1}{G}\sum_{i=1}^G R^{(i)},\qquad \hat\sigma_R := \sqrt{\frac{1}{G}\sum_{i=1}^G (R^{(i)}-\bar R)^2}.
\]

Every token in rollout $i$ receives the same trajectory-level advantage
$\hat A^{(i)}:=(R^{(i)}-\bar R)/\hat\sigma_R$.
Omitting clipping and KL terms, which affect optimization stability but not the population credit identity studied here, the estimator is
\begin{equation}\label{eq:grpo-grad-app}
\hat g_{\text{GRPO}} = \frac{1}{G}\sum_{i=1}^G \hat A^{(i)} \sum_{t=0}^{T-1} \nabla_\theta \log\pi_\theta(a_t^{(i)}\mid s_t^{(i)}).
\end{equation}

\mypar{Group rewards estimate the prompt value.}
\begin{lemma}[Baseline consistency]\label{lem:baseline}
The group statistics are Monte Carlo estimators of the value and its standard deviation:
\[
\E[\bar R]=p^\pi(s_0),\qquad \hat\sigma_R^2 \xrightarrow{\text{a.s.}} p^\pi(s_0)(1-p^\pi(s_0)) =: \sigma^2(s_0)\quad\text{as } G\to\infty.
\]
\end{lemma}

\begin{proof}
Each $R^{(i)}$ is an i.i.d.\ Bernoulli draw with mean $p^\pi(s_0)$; the result follows from the law of large numbers applied to the sample mean and sample variance.
\end{proof}

Under the large-$G$ approximation $\bar R\approx p^\pi(s_0)$ and $\hat\sigma_R\approx\sigma(s_0)$, the success-increment decomposition gives
\begin{equation}\label{eq:adv-as-sum-app}
\hat A^{(i)} \;\approx\; \frac{R^{(i)}-p^\pi(s_0)}{\sigma(s_0)} \;=\; \frac{1}{\sigma(s_0)}\sum_{t=0}^{T-1}\Delta p_t^{(i)}.
\end{equation}

Thus the trajectory-level GRPO advantage is the normalized sum of per-step success increments along that rollout. GRPO never observes the individual token-level increments, but the score-function identity recovers the same token-level credit in expectation.

\mypar{Recovering token-level credit.}
Under the same large-$G$ approximation,
\[
\E[\hat g_{\text{GRPO}}] \;\approx\; \frac{1}{\sigma(s_0)}\sum_{t=0}^{T-1}\E_\tau\!\left[(R(\tau)-p^\pi(s_0))\,\nabla_\theta\log\pi_\theta(a_t\mid s_t)\right].
\]

Fix a position $t$ and write $g_t=\nabla_\theta\log\pi_\theta(a_t\mid s_t)$, and the remaining expected terminal reward is
\[
\E[R(\tau)\mid s_t,a_t] = p^\pi(s_{t+1}).
\]

Therefore,
\[
\E_\tau[(R(\tau)-p^\pi(s_0))g_t]
= \E_\tau[(p^\pi(s_{t+1})-p^\pi(s_0))g_t].
\]

Decomposing $p^\pi(s_{t+1})-p^\pi(s_0)=\Delta p_t + p^\pi(s_t)-p^\pi(s_0)$ gives one desired term and one state-dependent baseline. The baseline term vanishes because
\[
\E_\tau[(p^\pi(s_t)-p^\pi(s_0))g_t]
= \E_\tau\!\left[(p^\pi(s_t)-p^\pi(s_0))\E_{a_t\sim\pi_\theta(\cdot\mid s_t)}[g_t\mid s_t]\right]=0,
\]

where $\E_{a\sim\pi_\theta(\cdot\mid s)}[\nabla_\theta\log\pi_\theta(a\mid s)]=\sum_a\nabla_\theta\pi_\theta(a\mid s)=\nabla_\theta 1=0$. Hence
\[
\E_\tau[(R(\tau)-p^\pi(s_0))g_t]
=\E_\tau[\Delta p_t\,g_t].
\]

Summing over $t$,
\[
\E[\hat g_{\text{GRPO}}] \;\approx\; \frac{1}{\sigma(s_0)}\,\E_\tau\!\left[\sum_{t=0}^{T-1}\Delta p_t\,\nabla_\theta\log\pi_\theta(a_t\mid s_t)\right] \;=\; \frac{1}{\sigma(s_0)}\,\nabla_\theta J(\pi_\theta),
\]

where the final equality is Proposition~\ref{prop:pg}. In the large-$G$ limit, GRPO therefore has the same population credit direction as the actor-critic policy gradient, up to the prompt-level normalization $1/\sigma(s_0)$. For finite groups, the sample mean and standard deviation introduce the usual Monte Carlo error of order $O(1/\sqrt{G})$.\hfill$\square$

\mypar{Implication.}
Proposition~\ref{prop:grpo} shows that the same mechanism applies to the algorithm used in our experiments, even though GRPO assigns only trajectory-level advantages and has no learned critic. Group rewards estimate the prompt-level success baseline, and the score-function identity recovers the same token-level success-increment credit in expectation. Thus changing the task structure can change which reasoning strategies RLVR reinforces: stricter tasks make high-bit-rate steps more reward-distinguishable, which explains why bijective or constrained variants shift the model toward derivation and improve OOD strategy generalization.

\subsection{Validation Accuracy Under Bit-Rate Interventions}
\label{sec:appendix_theory_intervention}

Sec.~\ref{sec:theory-verification} uses Short Budget and Constrained prompts to intervene the bit-rate prediction and validates our theoretical analysis; the central evidence is already in Fig.~\ref{fig:strategy-budgets}, where both interventions increase derivation across the OOD axes and directly confirm the theory's prediction about which reasoning strategy RLVR should reinforce. This appendix subsection is therefore mainly a completeness check: it reports the validation accuracy associated with those runs.

\begin{figure}[t]
  \centering
  \includegraphics[width=\linewidth]{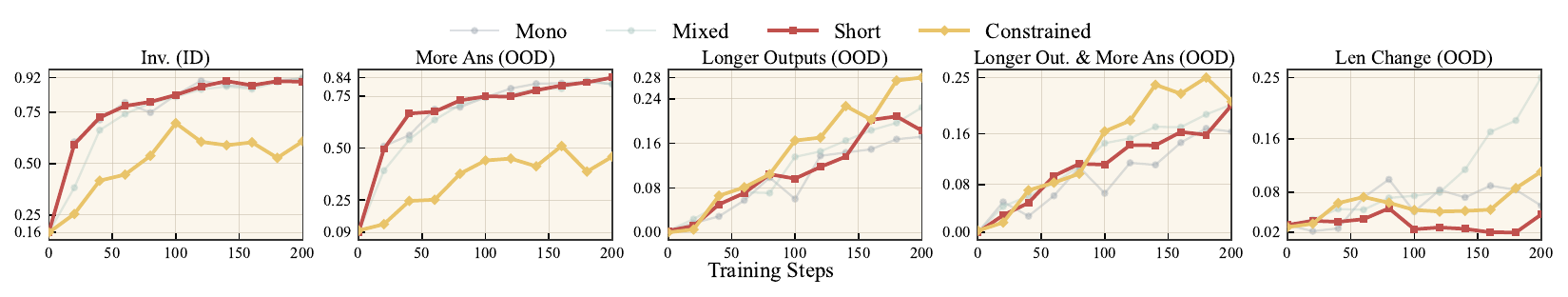}
  \vspace{-14pt}
  \caption{Accuracy under bit-rate interventions (Sec.~\ref{sec:theory-verification}).}
  \label{fig:constraints-budgets}
  \vspace{-4mm}
\end{figure}

Fig.~\ref{fig:constraints-budgets} plots the same five validation as Fig.~\ref{fig:strategy-budgets}. The two original baselines, \textsc{Mono} and \textsc{Mixed}, are shown in shallow colors for reference, while Short Budget and Constrained prompts are highlighted. The accuracy curves make an important qualification explicit: stronger derivation incentives do not by themselves guarantee the best final accuracy. The balanced \textsc{Mixed} run remains the strongest overall by final accuracy, especially on the most shifted \textsc{Len Change} axis, while Short Budget and Constrained prompts win only on some axes. This is consistent with the mixture-ratio analysis in Sec.~\ref{sec:varying_mixture} and the rollout audit in Sec.~\ref{sec:appendix_mix_errors}: reasoning strategy is a necessary driver of OOD generalization, but accuracy also depends on task-specific execution skills such as modular arithmetic, faithful code reading, and handling multi-valued inverses. Thus Fig.~\ref{fig:constraints-budgets} does not weaken the theory result in Fig.~\ref{fig:strategy-budgets}; it separates the strategy-selection mechanism from the additional execution reliability needed for the best end-to-end performance.

\section{Probing Details}
\label{sec:appendix_probing}

Sec.~\ref{sec:probing} asks when the hidden state contains information about the reasoning strategy that will appear in the generated trace. This section provides the detailed implementations and considerations.

\mypar{Setup.} We conduct probing for both \textsc{Mono} and \textsc{Mixed} models. For each checkpoint we evaluate four held-out OOD sets built from the same six non-bijective functions. Then, we extract the hidden states from the 21-th LLM layer (Qwen2.5-7B has 28 transformer layers) at the $N \in \{0, 100, 200, 300, 500, 1000, 1500\}$ positions after the $\texttt{<think>}$ token. This layer is intuitively containing the rich features and is not too close to the last layer simply mapping features to tokens.

In probing, we train the linear classifier to predict the derivation/enumeration labels annotated by our strategy categorization (Sec.~\ref{sec:appendix_strategy_categorization}). We use a logistic-regression probe with 5-fold stratified cross-validation. The folds preserve the evaluation set, function, and strategy label. Since the two strategies are imbalanced across models and evaluation sets, we report balanced accuracy, a commonly used metric for imbalanced classification that averages the recall of the two classes.

Specifically, let $z_i \in \{\mathrm{Derive}, \mathrm{Enum}\}$ be the strategy label for response $i$, and let $\hat z_i$ be the probe prediction. For a strategy class $c$, define
\[
\operatorname{recall}(c)
=
\frac{\sum_i \1[z_i=c \wedge \hat z_i=c]}
       {\sum_i \1[z_i=c]} .
\]
The balanced accuracy is
\[
\operatorname{BAcc}
=
\left(
\operatorname{recall}(\mathrm{Derive})
+
\operatorname{recall}(\mathrm{Enum})
\right) / 2.
\]

\mypar{Null Condition.} To validate that the representation indeed encodes strategy, instead of only relying on the shortcut of memorizing the statistics of function names and evaluation set properties, we construct a null condition as a contrast to determine the significance of linear probing. We adopt the shuffling of labels to create the null condition, and the labels are shuffled only among prompts that share the same evaluation set and function. This keeps the strategy frequency of every function on every evaluation set fixed, while breaking the alignment between individual hidden states and individual strategy labels. Table~\ref{tab:appendix_null_choice} shows why this is the appropriate and most rigorous null condition, compared with two less careful shuffling.

\begin{table}[t]
  \centering
  \small
  \setlength{\tabcolsep}{6pt}
  \caption{Comparison of null conditions. Shuffling at different levels lead to different rigor. We finally select the null condition that is the strongest, for the probing experiments (Sec.~\ref{sec:probing}).}
  \label{tab:appendix_null_choice}
  \begin{tabular}{llc}
    \toprule
    Null type (Shuffle) & What it preserves & Balanced Accuracy at $N=0$ \\
    \midrule
    Global    & Overall frequency            & 50\% \\
    Per Eval Set  & Per eval-set frequency & 65\% \\
    \textbf{Function $\times$ Eval (Ours)} & Both function and eval-set frequency & 76\% \\
    \bottomrule
  \end{tabular}
\end{table}

\mypar{Response Structure.} Another crucial detail in Sec.~\ref{sec:probing} and Fig.~\ref{fig:probe_strategy} is how we assign the semantic meanings to different positions inside the reasoning trace. Although the problems vary, with changes of how the model thinks, we find that the responses from the LLMs follow a similar structure, which is intuitively utilized to interpret the results in Fig.~\ref{fig:probe_strategy}. We demonstrate an illustrative example in Fig.~\ref{fig:appendix_probe_response_example}. Our token counts in Fig.~\ref{fig:probe_strategy} are estimated from manually inspecting 50 samples of derivation responses and 50 brute-force-enumeration responses per model, and we specifically calculate the position of the keywords that indicate the change of reasoning stage.

\begin{figure}[t]
\centering
\begin{tcblisting}{
  enhanced,
  colback=gray!5, colframe=gray!55,
  arc=2mm, boxrule=0.5pt,
  left=8pt, right=8pt, top=4pt, bottom=4pt,
  listing only,
  listing options={
    basicstyle=\ttfamily\tiny,
    breaklines=true,
    columns=fullflexible,
    keepspaces=true,
    showstringspaces=false,
  },
}
<think>
To find three different digit strings x such that main_solution(x) returns "22",
we need to analyze the function for inputs of length 2.

The function takes a string s of length 2 and computes the cyclic absolute
difference between each digit and the next digit. Let the input string be ab,
where a and b are digits. The function computes
  first output digit  = |a - b|
  second output digit = |b - a|.

We need the output to be "22", so both positions must equal 2:
  |a - b| = 2
  |b - a| = 2.

These constraints are symmetric, so it is enough to solve |a - b| = 2.

Case 1: a - b = 2.
  a=2, b=0 -> "20"
  a=3, b=1 -> "31"
  a=4, b=2 -> "42"
  a=5, b=3 -> "53"
  a=6, b=4 -> "64"
  a=7, b=5 -> "75"
  a=8, b=6 -> "86"
  a=9, b=7 -> "97"

Case 2: b - a = 2.
  a=0, b=2 -> "02"
  a=1, b=3 -> "13"
  a=2, b=4 -> "24"
  a=3, b=5 -> "35"
  a=4, b=6 -> "46"
  a=5, b=7 -> "57"
  a=6, b=8 -> "68"
  a=7, b=9 -> "79"

We only need three different inputs, so choose "20", "31", and "42".
Verification:
  "20" gives |2-0|=2 and |0-2|=2, so the output is "22".
  "31" gives |3-1|=2 and |1-3|=2, so the output is "22".
  "42" gives |4-2|=2 and |2-4|=2, so the output is "22".

All three inputs satisfy the condition.
</think><answer>20,31,42</answer>
\end{tcblisting}
\vspace{-3mm}
\caption{A representative derivation trace. The model first restates the question, then explains the function, and introduces variables to solve the problem.}
\label{fig:appendix_probe_response_example}
\vspace{-4mm}
\end{figure}

\section{Mixture Ratios: Detailed Error Analysis}
\label{sec:appendix_mix_errors}

Sec.~\ref{sec:varying_mixture} shows two facts that must be read together: adding more bijective data monotonically increases analytical derivation, but the balanced $1{:}1$ mixture is most accurate. To separate strategy choice from execution reliability, we sample 128 OOD prompts where the three ratio models, $1{:}3$, $1{:}1$, and $3{:}1$, disagree. For each prompt, we inspect the greedy response from all three ratios (384 traces total), and annotate the strategy and primary error type. Unless stated otherwise, the percentages are over all 128 audited prompts for that ratio.

Based on the contexts from Sec.~\ref{sec:varying_mixture}, we especially analyze the error case when bijective tasks is dominating (3:1), when it can correctly utilize the derivation strategy. (1) The $3{:}1$ model chooses derivation most often, but its execution-error rate rises to $32.8\%$ of all audited OOD samples ($24.2\%$ for $1{:}1$) and to $57.7\%$ on \textsc{Longer Outputs}. A representative failure comes from the held-out prefix-square rule. The code maintains a running state by \texttt{state = (state + digit) \% 10}, and each output digit is \texttt{(state * state) \% 10}. Equivalently, if $s_i$ is the running state after reading the $i$-th input digit, then the target digit must satisfy $y_i=s_i^2\bmod 10$. For target ``119641'', the model must therefore find input digits whose running states have square residues $1,1,9,6,4,1$ in order. All three ratios recognize this structure and begin solving the modular state constraints. The $3{:}1$ model, however, does not finish the chain: after working through only the early states, it writes that it will ``solve for $s_3, s_4, s_5$ similarly'' and then directly emits \texttt{102864} and \texttt{926844}. These are not valid completions. (2) The same model also outputs less solutions than required, covering $11.7\%$ cases. This also suggests the importance of refining the task-specific skills beyond meta-level reasoning strategies, as explained in Sec.~\ref{sec:varying_mixture}.

\section{Analysis of Zero-shot Math Evaluation}
\label{sec:appendix_math}

This section covers implementation details and disagreement examples supporting the analysis in Sec.~\ref{sec:math_zeroshot}. Sec.~\ref{sec:appendix_math_protocol} describes the evaluation protocol. The remaining subsections organize the qualitative analysis.

\subsection{Evaluation Protocol}
\label{sec:appendix_math_protocol}

We evaluate with the math-evaluation-harness~\citep{math_eval_harness}, using greedy decoding and the default prompt for a fair comparison across all the models. The benchmarks are MATH500~\citep{hendrycks2021math,lightman2023let}, Minerva-Math~\citep{lewkowycz2022minerva}, AIME24, AIME25, AMC23, and OlympiadBench~\citep{he2024olympiadbench}, scored with the harness's default evaluators.

\subsection{\textsc{Mixed} Faithful To Problem Structure}
\label{sec:appendix_math_structure}

As described in Sec.~\ref{sec:math_zeroshot}, the \textsc{Mixed} model can better keep the structure of the problem, which is consistent with the derivation strategy usage and our discover on the later strategy committment from hidden states.

\mypar{Example: AIME 2025 (base divisibility).}
\begin{quote}
\emph{Problem.} Find the sum of all integer bases $b > 9$ for which $17_b$ is a divisor of $97_b$.
\end{quote}
\noindent Here $97_b$ denotes the numeral $97$ written in base $b$, so its decimal value is $9b+7$; similarly, $17_b=b+7$. The notation $u\mid v$ means that $u$ divides $v$.
\textsc{Mixed} rewrites the divisibility condition by noting $97_b = 9b + 7$ and $17_b = b + 7$, then performs the algebraic identity $9b + 7 = 9(b+7) - 56$. Hence $b+7 \mid 56$. It enumerates the full divisor list $\{1, 2, 4, 7, 8, 14, 28, 56\}$, filters by $b > 9$ (equivalently $b+7 > 16$), and finds $b+7 \in \{28, 56\}$, i.e.\ $b = 21$ and $b = 49$. The answer is $21 + 49 = 70$ (correct). \textsc{Mono} reparameterizes the same condition as $9b + 7 = k(b+7)$ for some integer $k$, solves to $b = 7(k-1)/(9-k)$, and concludes that $9-k$ must divide $7$. It checks only the divisors $\{1, -1, 7, -7\}$ of $7$ instead of all divisors of $56$, finds $b = 49$ as the only base $> 9$, and outputs $49$. The mistake is not arithmetic: \textsc{Mono} collapses an enumeration over divisors of $56$ into one over divisors of $7$, dropping most of the case structure; \textsc{Mixed} keeps the full divisor set in scope.

\mypar{Example: AMC 2023 (cubic with integer roots).}
\begin{quote}
\emph{Problem.} For how many ordered pairs $(a, b)$ of integers does the polynomial $x^3 + a x^2 + b x + 6$ have three distinct integer roots?
\end{quote}
\noindent\textsc{Mixed} enumerates the eight ordered triples $(p, q, r)$ of distinct integers with $p q r = -6$, computes $(a, b) = (-(p+q+r),\, pq + qr + rp)$ for each, and deduplicates to the $5$ distinct pairs (correct). \textsc{Mono} lists only three such triples, misses $(-1, 1, 6)$, $(-2, 1, -3)$, $(2, -1, -3)$, and others, and outputs $2$.

\mypar{Example: AMC 2023 (custom operator).}
\begin{quote}
\emph{Problem.} For $u = a + bi$ and $v = c + di$, define $u \otimes v = ac + bdi$. Find $|z|^2$ given that $z \otimes z = z^2 + 40$.
\end{quote}
\noindent\textsc{Mixed} parses the operator correctly: for $z = x + yi$, $z \otimes z = x^2 + y^2 i$. It compares real and imaginary parts against $z^2 = (x^2 - y^2) + 2 x y i$ and solves to get $|z|^2 = 50$ (correct). \textsc{Mono} mis-substitutes into the operator definition, produces $z \otimes z = x^2 + y^2 i (x + y)$, and falls into a dead-end loop.

\subsection{\textsc{Mono} Better At Execution Reliability}
\label{sec:appendix_math_execution}

Alternatively, we consider the observation in Sec.~\ref{sec:varying_mixture} and Sec.~\ref{sec:math_zeroshot}, where the \textsc{Mono} model benefits from execution reliability, and has the advantage mostly from the computation-bottlenecked MATH500.

\mypar{Example: MATH500 (polynomial long division).}
\begin{quote}
\emph{Problem.} Find the quotient when $x^6 - 3$ is divided by $x + 1$.
\end{quote}
\noindent\textsc{Mixed} sets up polynomial long division correctly and performs the first two steps, then loses track mid-computation and outputs $0$. \textsc{Mono} carries the division through all six steps and obtains the correct quotient $x^5 - x^4 + x^3 - x^2 + x - 1$.

\mypar{Example: MATH500 (parallelogram fourth vertex).}
\begin{quote}
\emph{Problem.} A parallelogram has three of its vertices at $(5, 3)$, $(6, 8)$, $(7, 4)$, and the fourth at $(x, y)$ with $x > 7$. Find $x + y$.
\end{quote}
\noindent\textsc{Mixed} tries one diagonal pairing, finds $x = 6$ (which violates $x > 7$), and stops without trying alternatives. \textsc{Mono} also finds $x = 6$ from the first pairing, recognizes the constraint violation, re-evaluates with a different diagonal assignment, and finds $x = 8$, $y = 9$, giving $x + y = 17$ (correct).

\mypar{Example: MATH500 (group-size condition).}
\begin{quote}
\emph{Problem.} A class has between $30$ and $50$ students and cannot be broken into multiple groups of equal size. Adding two instructors to the count still does not allow such a split. How many students are in the class?
\end{quote}
\noindent\textsc{Mixed} misinterprets ``cannot be broken into multiple groups of equal size'' as ``not a perfect square'' and outputs $30$. \textsc{Mono} correctly interprets the condition as ``$n$ is prime,'' searches for twin primes in the range, and outputs $41$ (correct).

\section{Agentic Training Details}
\label{sec:appendix_agentic}

In Sec.~\ref{sec:agentic}, we explore applying our fix for reasoning strategies to agentic multi-turn tasks, and this section describes the detailed implementations. We follow RAGEN~\citep{wang2025ragen,wang2026ragen} and train the Qwen2.5-3B-Instruct model in the Sokoban environment. Since this task is deterministic, we can calculate the optimal step for each puzzle and utilize \emph{optimal\_steps + 2} actions to mark a stricter task than the original. We mix it with standard Sokoban at $1{:}1$ and compare against training on standard Sokoban alone, holding all other hyperparameters fixed.

In light of our theoretical analysis in Sec.~\ref{sec:theory}, we realize that the original reward design in Sokoban introduces large variances, \emph{e.g.}, the values for a total success is 10, and each box counts for a score 5, and no format reward guides the LLM to follow our action parser. Therefore, we add a 0.1 format reward to encourage following our thinking and answering format, then a 0.9 reward for pushing all the boxes on target, which is a more simple reward than the original Sokoban.

Same as our experiment in reasoning strategies, we use DAPO~\citep{dapo}, especially filtering out the groups without reward variances. As shown in Fig.~\ref{fig:sokoban_dapo}, the baseline DAPO run rises during the first 60 steps to a validation success rate of $24\%$, then drops sharply to below $10\%$ over the next 40 steps. With the stricter variant mixed in, training climbs steadily to roughly $30\%$ over the same window. The two curves track each other closely until step 60 and only diverge once the baseline begins to collapse, suggesting that the stricter signal enhances the agentic training. Although this experiment is not necessarily an OOD generalization, we hope this exploration could suggest the broader implication of our discovery and analysis for agentic scenarios in future works.

\section{Limitations, Societal Impacts, and Future Works}\label{sec:appendix_limitations}

\mypar{Limitations.} (1) Our study is intentionally controlled: the inversion suite makes reasoning strategies observable and OOD shifts clean, but it does not claim to exhaust all forms of reasoning in realistic tasks. In fact, we can extend the current inversion task into multiple string equations, the equations without invalid answers, etc., to investigate even more reasoning strategies of LLMs. (2) In addition, the current categorization focuses on analytical derivation versus brute-force enumeration because this is the decisive strategy contrast for inverse-function reasoning. More realistic scenarios may involve a mixture of strategies within the same trace, including decomposition, case analysis, local enumeration, self-verification, tool use, and planning, which would require a finer-grained strategy taxonomy. (3) Our theory also focuses on a clean search-MDP abstraction and identifies bits per token as one mechanism by which outcome rewards favor generalizable reasoning. A clearer theory could further model the latent function that maps task structure, strategy choice, and execution reliability to OOD success. (4) Finally, our agentic experiment is a small extension to hint the potential of our discovery for agentic tasks: it suggests that stricter task structure can stabilize Sokoban training, while extension to full-scale long-horizon agentic tasks would require better measures to create clean testbed of OOD generalization and reasoning strategies. (5) Our main analysis uses one base model, Qwen2.5-7B-Instruct, chosen as the operating point where the phenomenon is identifiable (Sec.~\ref{sec:appendix_base_models}). Other model families may sit at different operating points, and future analysis might need to select tasks localizing at the boundaries of such models' capabilities.

\mypar{Societal Impacts.} Our experiments inherit the general risks of LLMs, including possible biases from the base models, but this work mainly uses synthetic data and studies general reasoning behavior rather than user-facing deployment; if the identified task-structure principles improve OOD generalization more broadly, they may also reduce the need for collecting large amounts of human data and provide positive downstream societal impact.

\mypar{Future Works.}
(1) The most direct future direction is to extend the framework to richer reasoning patterns and more realistic scenarios where successful solutions naturally combine multiple strategies rather than selecting one dominant behavior. This would allow reasoning-strategy analysis to move from a clean binary distinction toward a compositional account of how models choose, combine, and switch strategies under distribution shift. (2) A second direction is to develop a sharper theory of strategy selection, especially by estimating strategy-dependent information gain from model traces, connecting it to learned representations, and characterizing when stricter auxiliary tasks improve strategy selection without sacrificing execution reliability. (3) A third direction is agentic RLVR. Our Sokoban result suggests that task structures demanding more useful information per action can matter beyond string inversion; future work can study how agents develop better reasoning patterns over long-term environments with memory, tools, exploration, and delayed rewards.

\end{document}